\documentclass[10pt,journal,compsoc]{IEEEtran}

\ifCLASSINFOpdf
\else
\fi

\usepackage{booktabs}
\usepackage{amsmath}
\usepackage{rotating}
\usepackage[pagebackref=true,breaklinks=true,
letterpaper=true,colorlinks,citecolor=blue,
linkcolor=blue,bookmarks=false]{hyperref}

\usepackage{color}
\usepackage{times}
\usepackage{epsfig}
\usepackage{graphicx}
\usepackage{amssymb}
\usepackage{multirow}
\usepackage{algorithm}
\usepackage{algorithmic}
\usepackage{array}
\usepackage{cite}
\newcommand{\RN}[1]{%
  \textup{\uppercase\expandafter{\romannumeral#1}}%
}
\DeclareMathOperator*{\argmax}{argmax}

\begin{document}

\title{Detecting and Tracking of Multiple Mice Using Part Proposal Networks}

\author{Zheheng Jiang, Zhihua Liu, Long Chen, Lei Tong, Xiangrong Zhang, Xiangyuan Lan, Danny Crookes \IEEEmembership{Senior Member,~IEEE}, Ming-Hsuan Yang \IEEEmembership{Fellow,~IEEE} and Huiyu Zhou
\thanks{Z. Jiang is with School of Computing and Communication, Lancaster University, United Kingdom. E-mail: z.jiang01@lancaster.ac.uk.}
\thanks{Z. Liu, L. Chen, L. Tong and H. Zhou are with School of Computing and Mathematical Sciences, University of Leicester, United Kingdom. E-mail: \{zl208;lc408;lt228;hz143\}@leicester.ac.uk. H. Zhou is the corresponding author.}
\thanks{X. Zhang is The Key Laboratory of Intelligent
Perception and Image Understanding of Ministry of Education, Xidian University, China. E-mail: xrzhang@ieee.org.}
\thanks{X. Lan is with the Shenzhen NS-Tech Co.Ltd., China. E-mail: yuanlan@life.hkbu.edu.hk.}
\thanks{D. Crookes is with School of Electronics, Electrical Engineering and Computer Science, Queen's University Belfast, United Kingdom. E-mail: d.crookes@qub.ac.uk.}
\thanks{M.-H. Yang s with the School of Engineering, University of California at
Merced, Merced, CA. E-mail: mhyang@ucmerced.edu.}

\thanks{Manuscript submitted in Apr 2020; revised xxxx.}}

\maketitle

\begin{abstract}
The study of mouse social behaviours has been increasingly undertaken in neuroscience research. However, automated quantification of mouse behaviours from the videos of interacting mice is still a challenging problem, where object tracking plays a key role in locating mice in their living spaces. Artificial markers are often applied for multiple mice tracking, which are intrusive and consequently interfere with the movements of mice in a dynamic environment. In this paper, we propose a novel method to continuously track several mice and individual parts without requiring any specific tagging. Firstly, we propose an efficient and robust deep learning based mouse part detection scheme to generate part candidates. Subsequently, we propose a novel Bayesian-inference Integer Linear Programming Model that jointly assigns the part candidates to individual targets with necessary geometric constraints whilst establishing pair-wise association between the detected parts. There is no publicly available dataset in the research community that provides a quantitative test-bed for the part detection and tracking of multiple mice, and we here introduce a new challenging \textit{Multi-Mice PartsTrack} dataset that is made of complex behaviours. Finally, we evaluate our proposed approach against several baselines on our new datasets, where the results show that our method outperforms the other state-of-the-art approaches in terms of accuracy. We also demonstrate the generalization ability of the proposed approach on tracking zebra and locust.
\end{abstract}


\IEEEpeerreviewmaketitle

\section{Introduction}
\IEEEPARstart{I}{n} neuroscience research, animal models are valuable tools to understand the pathology and development of neurological conditions such as Alzheimer’s and Parkinson's diseases~\cite{liu2017modeling,vogel2013neuron,
kalueff2016neurobiology}. Visual tracking of animals \cite{wang2019robust,villacorta2013neural,araujo2014self} is an essential task for many applications and has been successfully used in mice behavior analysis \cite{de2012computerized}. Scientific experiments in laboratories with mice need long-term observations by researchers and other parties. However, manually annotating long video recordings is a time-consuming task. Furthermore, manual documentation  suffers from a number of limitations such as being highly subjective and having scarce replicability. Hence, there is an increasing interest in the development of systems for automated analysis of mice social behaviour from videos\cite{jhuang2010automated,jiang2018context,burgos2012social}. 

Definitions of social behaviours can vary to some extent, from all the behaviours that occur when two or more animals are present in the scene \cite{altmann1974observational}, to only those behaviours in which one influences another \cite{sokolowski2010social}. Despite its definition, to automatically analyse social behaviours, discriminative features are often required: to record behaviours, to track the positions (or parts) of the participants, to identify individuals across time and space, and to quantify animal interactions. For automated behaviour analysis, a reliable and smart tracking method is required to associate the detected behaviours with the correct individuals. Moreover, accurate locations of mouse parts obtained from a stable tracking method enable the representation of interactions and allow for behavioural classification. Many applications, such as \cite{spink2001ethovision,giancardo2013automatic,galsworthy2005comparison}, analyse mouse behaviours using the tracking results of mice. Simultaneous tracking of two or more individuals poses a challenge in the computer vision community. The fact is that mice are mostly identical and highly deformable objects. In addition, social interaction between individuals makes the tracking mission even more complicated due to frequent occlusions. A popular method to track individuals during interactions is to label each subject with a unique marker, e.g., by bleaching~\cite{ohayon2013automated}, color~\cite{ballesta2014real}. Also, Galsworthy et. al \cite{galsworthy2005comparison} monitor multiple mice in a single cage by using radio transmitters buried under the skin and then record their activities by the detection coils. However, these systems are invasive for the tested subjects, and the labelling method in these systems very likely influences an individual's social behaviours as it frequently provides an olfactory and/or visual stimulus~\cite{dennis2008appearance}. Therefore, people generally prefer markerless identification.

\begin{figure*}
\begin{center}
\includegraphics[width=14cm]{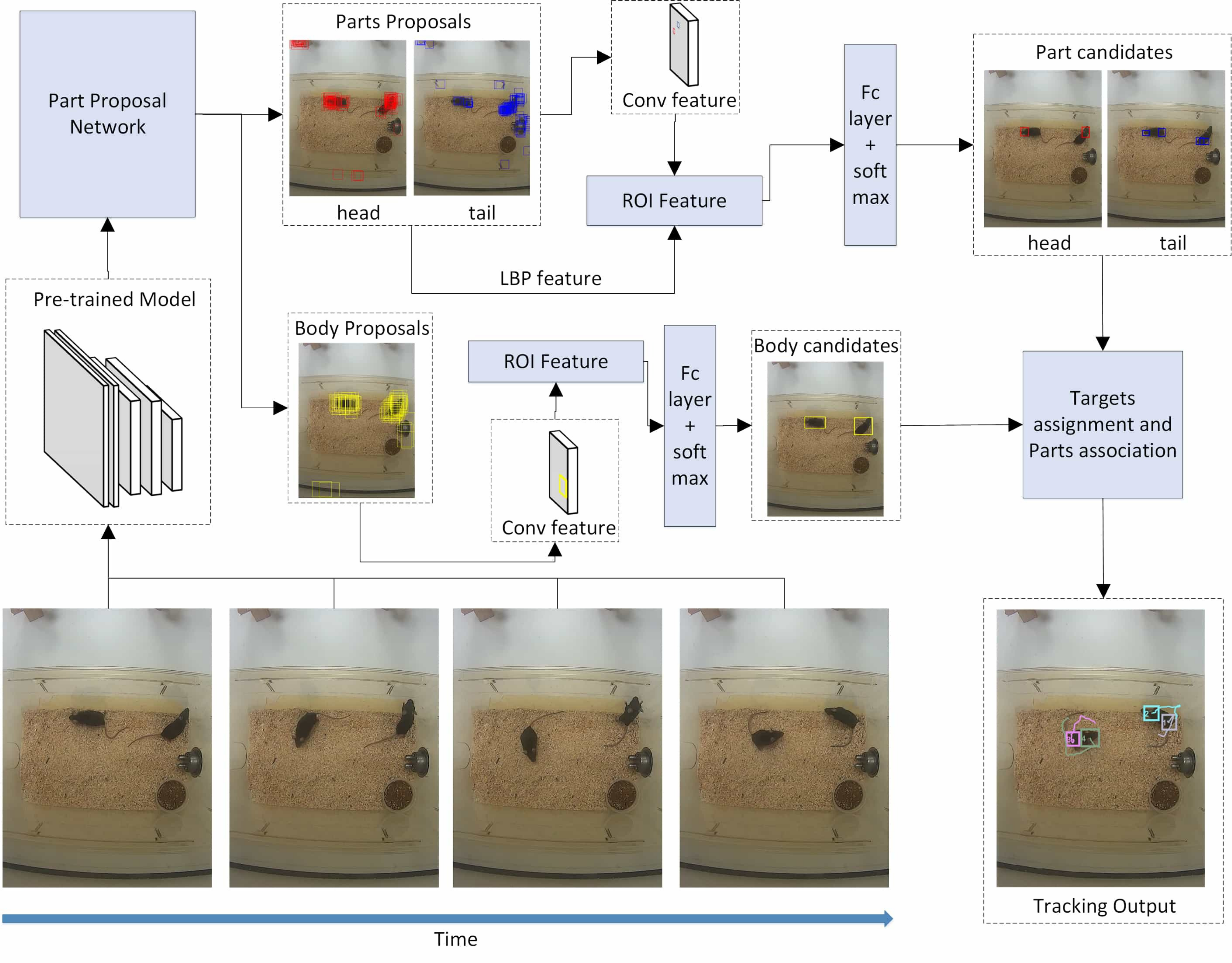}
\end{center}
\caption{Architecture of the proposed tracking system. Firstly, we propose an efficient and robust deep learning based mouse part detection scheme, which consists of a multi-stage Part Proposal Network for generating high-quality part and body proposals and two network branches for generating part and body candidates. Subsequently, we propose a novel Bayesian-Inference Integer Linear Programming Model that jointly assigns the part candidates to individual targets with necessary geometric constraints whilst establishing pair-wise association between the detected parts.}
\label{fig:tracking}
\end{figure*}

Several markerless approaches have been proposed to tackle this challenging problem. A common approach is to count on suitable foreground segmentation, which separates the extracted foreground pixels into several spatially connected groups using clustering algorithms such as Expectation-Maximization for Gaussian Mixture Models (e.g. \hspace{1sp}\cite{branson2005tracking,
ohayon2013automated,perez2014idtracker}) or watershed segmentation (e.g. \hspace{1sp}\cite{giancardo2013automatic}). However, if two mice are spatially close, in contact or occluding each other, it can be difficult to separate them only based on foreground detection. Moreover, the location estimation of mouse parts is unreliable when occlusion occurs. De Chaumont et al. \cite{de2012computerized} make use of  prior knowledge when tracking the mouse parts, and developed a model that connects a set of geometrical primitives under physics-based principles and adapts the model's parts to the moving mouse based on the defined physical engines. However, this type of  methods requires sophisticated skeleton models which are hard-coded in the system, and thus limit the flexibility of the methods.

The above methods belong to the family of detection-free tracking methods, which requires manual initialization of the mouse location in the first frame, and then tracks mice and their parts in subsequent frames. Although these methods have no need of pre-trained object detectors, they are prone to drifts and identity switches. Recently, tracking-by-detection methods have become popular as they are efficient at handling deformation and occlusion. The essential idea is to first detect objects and then handle the data association problem over frames. This approach has been widely used for human tracking and pose estimation~\cite{pishchulin2016deepcut,insafutdinov2016deepercut,zheng2018spatially,tang2015subgraph,liu2014learning}. Most recently, Wang et al. \cite{wang2019robust} propose a motion detection based system to monitor small target such as insects.

To address the part tracking problem of multiple mice in the context of tracking-by-detection, we break the original problem into three interdependent but also associated sub-problems. First, the positions of all the parts in each frame must be identified. Second, the detected parts must be assembled to form a physical mouse. Finally, the positions of the parts must be connected across image frames in order to generate mouse motion trajectories.

Detecting mouse parts in the first sub-problem is challenging due to the small size and subtle local inter-class differences between image frames. To address this problem, we propose a mouse parts and body detection framework based on the features from deep neural networks. In fact, the use of deep neural networks has already obtained promising outcomes for part detection and pose estimation of humans in some challenging benchmarks~\cite{pishchulin2016deepcut,insafutdinov2016deepercut,insafutdinov2017arttrack}. This suggests that deep learning architectures can be applied to part detection of lab animals.
For the second and third problems, we propose a Bayesian Inter Linear Program (BILP) model that resolves these problems by minimizing a joint objective function through Bayesian Inference. We wish to handle targets-candidates assignment and pair-wise part association within a single cost function. In order to solve the challenging problems in mouse part detection and tracking, we here design and evaluate a novel framework based on part association. Figure \ref{fig:tracking} shows the flowchart of the proposed tracking system. In summary, we have the following novel contributions:

1. We propose a reliable detection framework that is efficient for identifying mouse body and parts and obtains competitive performance on detection benchmarks.

2. With the detection candidates generated by the proposed detector, a {Bayesian-Inference Integer Linear Program Model} is proposed to estimate the locations of all the mouse parts present in an image. The formulation is based on the targets' assignment and pair-wise part association, subject to mutual consistency and exclusion constraints.

3. We formulate the parts' localization as a Bayesian inference process that combines the output of the proposed detector with prior geometric models of mice. Unlike the previous work such as \cite{de2012computerized}, our prior on the configuration of the mouse parts is not hard-coded but derived from our rich collection of the labeled training samples. In addition to geometric models, we also introduce motion cues to compensate for the missing appearance information.

4. Since there is no publicly available dataset that provides a foundation to quantitatively evaluate the Multi-Mice part tracking, we also introduce a new challenging \textit{Multi-Mice PartsTrack} dataset that comprises a set of video recordings of two or three mice in a home cage with the top view. Several common behaviours are included in this dataset such as `approaching', `following',`moving away',`nose contact',`solitary' and `Pinning'. The true part and body locations of mice were manually labeled for each image frame.

\section{Related Work}
In this section, we review the established approaches related to our proposed system. Section 2.1 reviews the existing methods for tracking multiple mice. Section 2.2 discusses the methods used for multi-people tracking.

\subsection{Mouse based Tracking}
Typically, the separation of foreground and background can be used as the first step of a multi-subject tracking algorithm, e.g. \cite{branson2005tracking,ohayon2013automated,perez2014idtracker,giancardo2013automatic,de2012computerized,twining2001robust,pistori2010mice,hong2015automated}. In mouse tracking, the knowledge of the foreground can be used to improve the accuracy of the tracking scheme.

Some existing approaches for tracking multiple mice focus on modelling the appearance of a mouse. For example, Hong et al.\cite{hong2015automated} first apply background subtraction and image segmentation from the top-view camera. They then fit an ellipse to each mouse in the foreground. Thus, the position and body orientation of each mouse are described by the fitted ellipse. Twining et al.\cite{twining2001robust} find mice in a new image by fitting active shape models to different locations in the image and selecting the instance that best fits the models. Similarly, de Chaumont et al.~\cite{de2012computerized} manually define shape models based on geometrical primitives and fit these models to images using a physical engine. Although these methods are fast and work well, the flexibility of such methods is limited as they require sophisticated skeleton models. Moreover, if two mice are touching, or one mouse is occluding the other, it can be difficult to separate them based only on the shapes of the blobs of the foreground pixels. Alternatively, P{\'e}rez-Escudero et al. \cite{perez2014idtracker} use fingerprints extracted from image frames in which the mice are not interacting. The fingerprints are then used to deal with occlusions and identity shifts. The idtracker developed by Romero-Ferrero et al.\cite{romero2019idtracker} identifies animals that cross certain paths using deep-learning-based image classifiers when tracking multiple animals.

In general, motion is a useful cue for dealing with occlusion. Model-based tracking approaches can benefit the identification by incorporating motion cues into the used model. A frequently adopted tracking method that combines both appearance and motion is based on particle filtering. For example, Pistori et al. \cite{pistori2010mice} extend a particle filtering approach with certain variations on the observation model to track multiple mice from the top view. Branson et al.\cite{branson2005tracking} present a particle filtering algorithm by tracking the contours of multiple mice and acquire the images of the targets from a side view. The ability of a particle filter to correctly track mice depends on how well the observation model works with mice. Due to the highly deformable shapes of mice, it is very difficult to explicitly model the entire mouse.

Another solution towards the problem of occlusions is to mark mice. Shemesh et al.\cite{shemesh2013high}, for example, apply fluorescent paints that light up in different colors under UVA light, and Ohayon et al.\cite{ohayon2013automated} dye the mouse fur with different patterns of strokes and dots. An alternative to visible markers are radio-frequency identifiers (RFID), which are used by Weissbrod et al.\cite{weissbrod2013automated} to identify individuals in combination with video data. However, these markers must be placed before recording and may be a potential distraction to mice. To deal with these issues, Giancardo et al.\cite{giancardo2013automatic} use a high-resolution thermal camera to detect minor differences in body temperature. 3D tracking can also disambiguate occlusions by using depth cameras (e.g. Hong et al.\cite{hong2015automated}) or multiple video cameras at different viewpoints (e.g. Sheets et al. \cite{sheets2013quantitative}), which is a  challenging solution requiring additional equipment and calibration work. The summary of mouse based tracking is shown in Supplementary A.

\subsection{People based Tracking}
Human tracking is also a well studied topic in computer vision. Some of the techniques developed for human tracking may be applied for mouse tracking. Most filter-based tracking methods such as Kalman filter, Particle filter and correlation filter are well suited for online applications due to their recursive nature. Some approaches of this category focus on tracking a single target by model evolution \cite{zhang2017multi,zheng2019dynamically}. Some of the others aim at training better object classifiers \cite{ma2017robust} or learning better target representations \cite{zheng2018spatially}. However, these methods cannot guarantee a global optimum as they conduct tracking on each target individually.

To alleviate the problems of filter-based tracking methods, tracking-by-detection methods have been used in many applications. The fundamental idea is to first detect objects in each frame and then address the data association problem. Recent approaches in this category have been focused on improving the performance of the developed object detectors or designing better data association techniques to improve tracking performance. For example, Shu et al. in\cite{shu2012part} propose an extension to deformable part-based human detector \cite{felzenszwalb2010object} and utilize the visible part to infer the state of the whole object. A number of approaches rely on data association methods such as the Hungarian algorithm \cite{xiang2015learning,bewley2016simple}, global objective function optimization\cite{milan2013continuous,zhang2008global,pirsiavash2011globally,shitrit2014multi,hamid2015joint} and Recurrent Neural Network \cite{milan2017online}. Xiang et al. \cite{xiang2015learning} propose to formulate tracking as a Markov decision process with a policy learned on the labelled training data. They aim to learn similarity scores between targets and detections, which are used in the Hungarian algorithm to obtain the targets' assignment. In \cite{milan2013continuous}, the authors propose a continuous formulation that analytically models mutual occlusions, dynamics and trajectory continuity, but utilizes a simple SVM detector in their appearance model. Rezatofighi et al. \cite{hamid2015joint} propose an efficient approximation of Joint Probabilistic Data Association (JPDA) to reduce computational complexity in data association. Recent approaches have formulated data association as a network
flow problem and then seek the solution by optimizing a global objective function. Zhang et al. \cite{zhang2008global} show that a promising solution towards the network flow problem can be found in polynomial time and is highly efficient in practice. Pirsiavash et al. in \cite{pirsiavash2011globally} also adopt network flows, and use dynamic programming to search for a high quality sub-optimal solution. Authors in \cite{shitrit2014multi} propose a multi-commodity network flow to incorporate the appearance consistency between the groups of people. Milan et al. \cite{milan2017online} propose data-driven approximations to deal with the data association problem using long short-term memory (LSTM) networks whilst utilizing RNNs for targets' prediction and updating. Compared against these methods, we consider targets' assignment and pair-wise part association in our problem formulation. Also, we utilize Bayesian Inference to construct correct correspondences between the proposals and the measurements. Moreover, we introduce a geometric constraint in Bayesian Inference to help reduce the ambiguity caused by the nearby targets with similar appearance.

\subsection{Pose Estimation}
Recent methods for multi-person pose estimation can be broadly divided into two categories, i.e., top-down and bottom-up. Top-down methods \cite{pishchulin2012articulated,gkioxari2014using,he2017mask} aim to first detect people in the image and then estimate body pose independently. For example, Pishchulin et al. \cite{pishchulin2012articulated} follow this paradigm by incorporating a trained deformable part model into a pictorial structure framework. Gkioxari et al \cite{gkioxari2014using} use a person detector modelled after poselets, which is more robust to occlusions. The very recent Mask-RCNN method \cite{he2017mask} extends the standard Faster R-CNN by adding a branch for predicting object masks, which shows  satisfactory results in person keypoint estimation. However, such approaches are applicable only if people appear well separable. Moreover, these pose estimation methods always output a fixed number of body joints without accounting for occlusion. Bottom-up methods firstly detect body parts instead of full persons, and then associate these parts to individual instances. Pishchulin et al. \cite{pishchulin2016deepcut}, and later Insafutdinov et al. \cite{insafutdinov2016deepercut} generate body parts hypotheses using CNN-based part detectors, and then jointly solve the problems of people detection and pose estimation by formulating these problems as part grouping and labelling via an integer linear program. Cao et al. \cite{cao2016realtime} combine the unary joint detector modified from \cite{wei2016convolutional} with a part affinity field and greedily generate person instance proposals. In view of the ability of bottom-up approaches, they are more robust to occlusion as they assemble parts to the corresponding individuals. However, these methods do not incorporate any part tracking and therefore only work on single images.

In the case of mice, previous studies commonly use the top-down scheme. For example, Cootes et al.~\cite{cootes1995active} and Twining et al.\cite{twining2001robust} find mice in a testing image by fitting active shape models to different locations in the image and selecting the instance that best fits the models. Similarly, de Chaumont et al.~\cite{de2012computerized} manually define shape models based on geometrical primitives and fit these models to the image using a physical engine. Although these methods work effectively, the flexibility of such methods is limited due to the requirement of sophisticated skeleton models. Most recently, based on previous designs of neural networks for human pose estimation, Mathis et al. \cite{mathis2018deeplabcut} develop their pose estimation model by deploying transfer learning that can reduce the number of the required training examples. Pereira et al. \cite{pereira2019fast} incorporate the previous work with a graphical user interface for labeling of body parts whilst training the network. However, these animal pose estimation systems are based on the single-image pose estimation methods.
\section{Proposed Mouse parts and body Detection}
As stated before, a powerful part and body detector supplies the solid foundation for successful body tracking. Driven by the recent development in deep neural networks with `attention' mechanisms, we here propose a novel mouse part and body detection framework which consists of two components: a multi-stage Part Proposal Network that generates suitable proposals of mouse parts and body, followed by a fully connected network that classifies these proposals using higher-resolution convolutional and Local Binary Pattern (LBP) features extracted from the original images.

\subsection{Part Proposal Network (PPN) for Mouse parts and body Detection}
\begin{figure}
\begin{center}
\includegraphics[width=9cm]{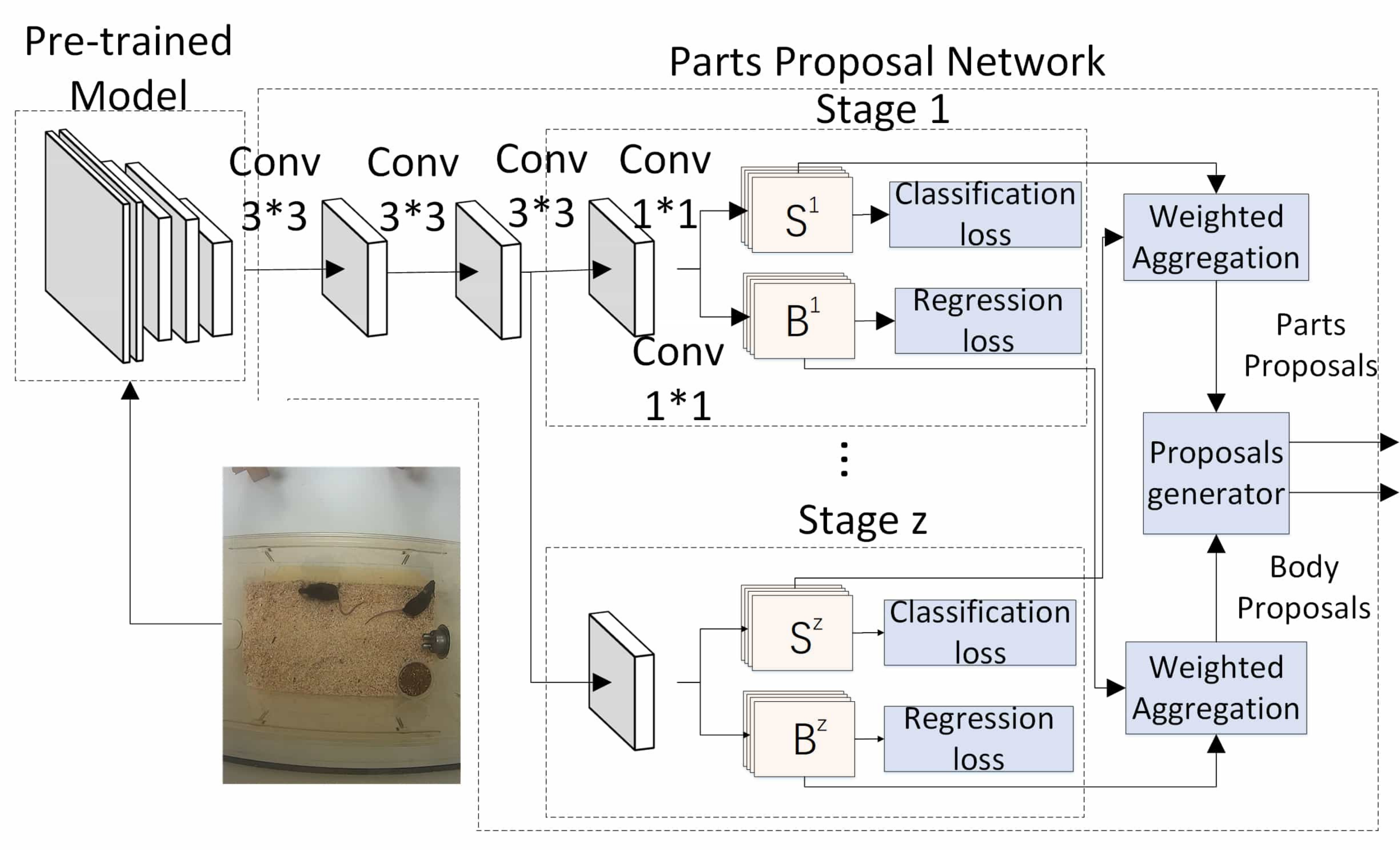}
\end{center}
\caption{Architecture of the proposed Part Proposal Network. A multi-stage architecture is adopted to perform effective hard negative mining for generating high-quality object proposals. The final confidence map and bounding boxes are attained by aggregating the weighted output of each stage.}
\label{fig:training}
\end{figure}

Detecting mouse parts is non-trivial due to their small size and subtle local inter-class differences across images. To develop a strong detector, we firstly create a proposal generation network which directs the downstream classifier where to look. The region proposal network (RPN) in the Faster R-CNN~\cite{ren2015faster} and selective search \cite{uijlings2013selective,girshick2015fast} are two state-of-the-art methods for object proposal generation. These methods were developed as a class-agnostic proposal generator aiming to create the bounding-boxes that may contain objects. Without specific knowledge, these region proposals may not be accurate. Deformable Part Model (DPM) \cite{felzenszwalb2010object} is another frequently used algorithm to generate small object proposals. DPM and its variants are a class-specific method, which is based on the Histogram of Oriented Gradient (HOG) features. In addition, most negative examples generated by these methods belong to negative examples, because they are randomly selected from the background. Unlike these methods, we adopt a multi-stage architecture to perform effective hard negative mining. Suppose we can have finer region proposals, and the accuracy of the proposed classifier can be further improved. In this section, we describe the proposed multi-stage PPN as shown in Fig. \ref{fig:training}, which can generate high-quality object proposals from the convolutional feature maps.

\begin{figure}
\begin{center}
\includegraphics[width=8.5cm]{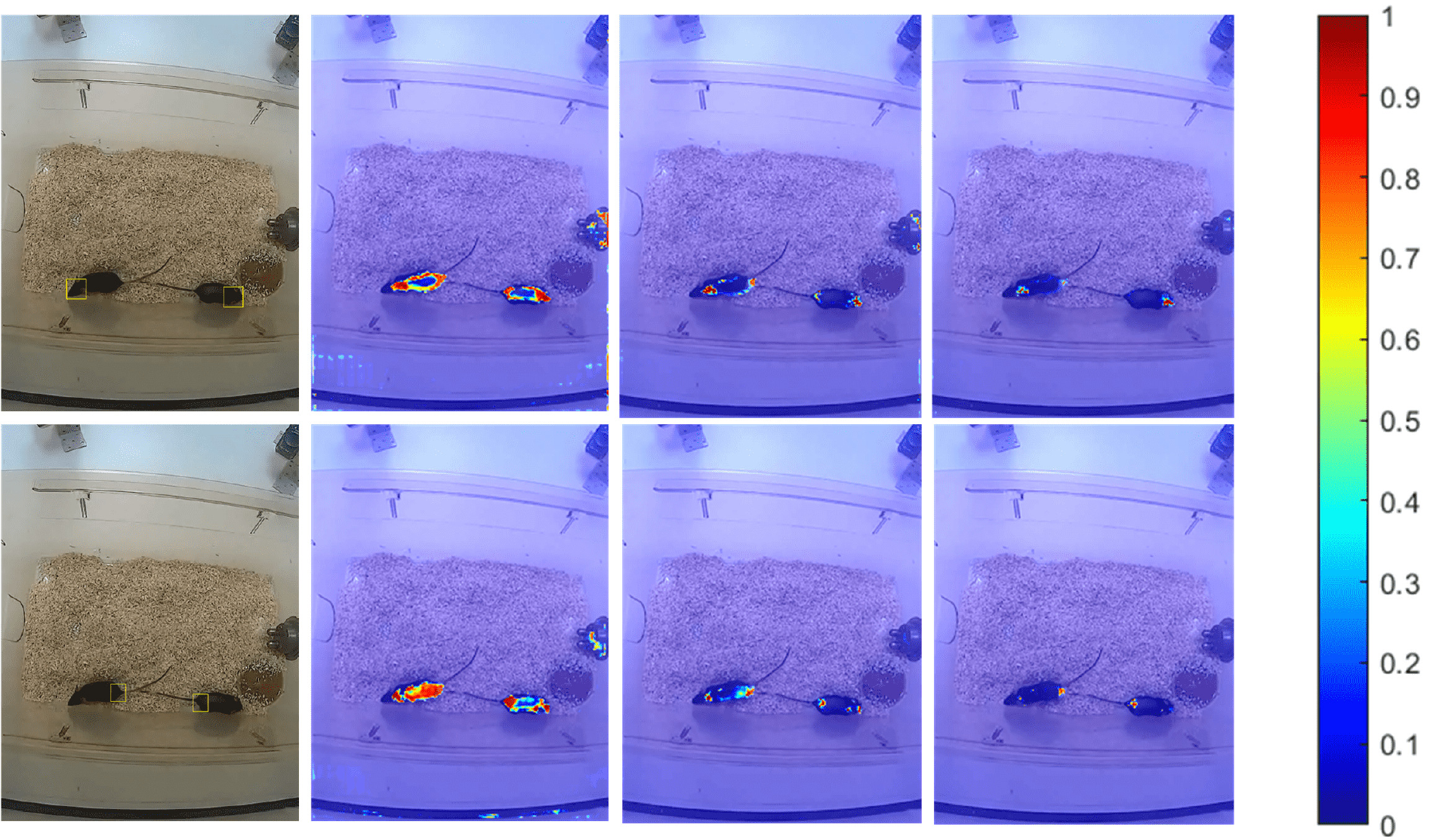}\\\qquad Stage 1 \qquad Stage 2\qquad Stage 3
\end{center}
\caption{Confidence maps of the root boxes of the mouse head (first row) and tail base (second row). More intense shades of red indicate a higher confidence score and lighter shades of blue indicate a lower confidence score. In stage 1, there is a confusion between the mouse head and the tail, as several parts of mouse have high confidence scores. Moreover, some of the high confidence scores come from the background. However, the estimates are increasingly refined in the later stages.}
\label{fig:hard_negative}
\end{figure}

We firstly adopt the VGG-19 net~\cite{simonyan2014very} as the backbone network, which is pre-trained on the ImageNet dataset~\cite{russakovsky2015imagenet}. The input image is first analysed by a convolutional network initialized by the first 15 layers of the VGG-19~ net\cite{simonyan2014very} and fine-tuned for the region proposal task to generate a set of feature maps as the input to our proposed PPN. We use a partial VGG-19 network and add another two trainable convolutional networks after the VGG-19 network, aiming to extract discriminative features and generate feature maps $F$ for the subsequent proposal generation tasks. To produce the region proposals, we construct a multi-stage class-specific proposal generation network. In each stage, we create a new network including an intermediate 3*3 convolutional layer sliding on feature maps $F$, followed by two sibling 1*1 convolutional layers for classification and the bounding box regression respectively. For each sliding window in the 3*3 convolutional layer, we simultaneously predict multiple region proposals (called root boxes) in the original image. In our method, the multiple scales and aspect ratios of the root boxes are different for individual objects, dependent on the object size in the image. The classification layer provides the confidence scores of the root boxes, as used to mine and add negative root boxes to a mini-batch of the next stage. Each mini-batch contains several positive and negative example root boxes from the image. Initially, all the positive root boxes in the image are selected and the negative root boxes are randomly sampled. In each stage, hard negative root boxes are used to replace the previous negative root boxes in the mini-batch. Here, we define $\mathcal{D}_{pos}^{z}\subset \mathcal{D}^{z}$ and $\mathcal{D}_{neg}^{z}\subset \mathcal{D}^{z}$, consisting of all the positive and negative root boxes respectively in a mini-batch at stage $z$. $\mathcal{D}^{z}$ is a universal set containing all the positive, negative and unlabelled root boxes in the image.

With the above definitions, we infer a multi-task loss function for the image at stage $z$ as follows:
\begin{equation}
\begin{split}
L^{z}=&-\frac{1}{\mathcal{N}}\sum_{j}\left (\sum_{i\in \mathcal{D}_{pos}^{z}}\log S_{j}^{z} \left ( i \right )+ \right.\\&\left. \sum_{i\in \mathcal{D}_{neg}^{z}}\log \overline{S}_{j}^{z} \left ( i \right ) +   \sum_{i\in \mathcal{D}_{pos}^{z}} L_{reg}\left (B_{j}^{z} \left ( i\right ) - \hat{B}_{j}^{z}\left ( i\right )\right )\right )
\label{equation:loss}
\end{split}
\end{equation}
Here, $S_{j}^{z} \left ( i \right )$ denote the predicted probability of the root box $i$ being a positive object of the mouse body $p_{0}$ or a mouse part $p_{j}\in \left\{ p\right\}_{j=1}^J$ at stage $t$. The negative predicted probabilities are represented as $\overline{S}$. For the regression loss $L_{reg}$, we use the robust loss function (smooth $L1$) defined in  \cite{ren2015faster}. $B_{j}^{z}\left ( i \right )$ is a vector which represents the 4 parameterized coordinates of the predicted bounding box associated with the root box $i$ of $p_{j}$ at stage $z$, and $\hat{B}_{j}^{z}$ is that of the ground-truthed box. All the terms are finally normalized by the mini-batch size (i.e., $\mathcal{N} = 256$).

In order to utilize our region proposal networks trained at each stage, we assign the weights to the output of the classification-regression networks (represented by $E^{z}$) based on their pseudo-loss:
\begin{equation}
\begin{split}
E^{z}=\frac{1}{num(\mathcal{D}^{z})*j}\sum_{j}\left (\sum_{i\in \left (\mathcal{D}^{z}-\mathcal{D}_{pos}^{z}\right )} S_{j}^{z} \left ( i \right )+\sum_{i\in \mathcal{D}_{pos}^{z}} \overline{S}_{j}^{z} \left ( i  \right )\right )
\label{equation:pseudo_loss}
\end{split}
\end{equation}
\begin{equation}
\begin{split}
\alpha^{z}=\frac{1-E^{z}}{\sum_{z=1}^{Z}\left(1-E^{z}\right )}
\label{equation:RPNs_weight}
\end{split}
\end{equation}
where $num$ is a function to count the number of the root boxes. After $Z$ stages of the training, the confidence scores $\mathcal{S}_{j}$ and the bounding boxes $\mathcal{B}_{j}$ of the root boxes with respect to object $p_{j}$ have the following form: $\mathcal{S}_{j}=\sum_{z}\alpha^{z}S_{j}^{z}$ and $\mathcal{B}_{j}=\sum_{z}\alpha^{z}B_{j}^{z}$.

Fig. \ref{fig:hard_negative} shows the refinement of the confidence maps across different stages. To reduce the number of the proposals for efficiency, we remove all the bounding boxes that have an intersection-over-union (IoU) \cite{ren2015faster} ratio over 0.7 with another bounding box that has a higher detection confidence. Supplementary B summarises the training steps of our PPN.

\subsection{Part candidates' generation}

With the proposals generated by PPN, inspired by the Faster R-CNN layer~\cite{ren2015faster}, we also crop the proposals and extract fixed-length features from the feature maps. However, the feature maps of the faster R-CNN are of low resolution for detecting small objects (e.g. mouse heads and tails). Generally, a small proposal box is mapped onto only a small region (sometimes 1*1*$n$) at the last pooling layer. Such small feature map lacks discriminative information, and thus degrades the following classifier. We address this problem using the pooling features from the shallower layers, and by additionally extracting texture features (e.g. LBP features) from the original image. For classifier training, we construct the training set by selecting the top-ranked 200 proposals (and ground truths) of each image for each class. At the testing stage, we use a trained classifier to classify all the proposals in an image, and then perform non-maximum suppression independently for each class to obtain part candidates.
\section{Parts Tracking of Multi-mouse}
In order to achieve successful body and part tracking for all the mice freely interacting in a home cage, we here propose a Bayesian-inference Integer Linear Program (BILP) formulation. Solving this BILP will yield an optimal solution to provide the locations of mouse parts (i.e., head and tail) in continuous videos, whilst fulfilling mouse part association over time. Then the mouse body is obtained by connecting tracked mouse parts.

\subsection{Problem Formulation}
Given a video sequence $\mathcal{F}$ containing $N^{*}=\left \{ 1,...,N \right \}$ parts, we generate a set of detection candidates $M^{*}_{t}=\left \{ 1,...,M_{t} \right \}$ using the proposed detector at time $t$. For the task of multi-mouse part tracking, our goal is to jointly address two problems: (1) Successful assembling of the detected parts to represent each individual mouse, and (2) correct alignment of the mouse parts to form motion trajectories.

For the benefit of representation, similar to \cite{pishchulin2016deepcut,insafutdinov2016deepercut} we define two binary vector spaces $\mathbb{B}^{M_{t}*M_{t}}$ and $\mathbb{B}^{N*\left ( M_{t}+1 \right )}$, and encode these two problems through the two binary vectors of $\gamma_{t}\in\Gamma_{t}\subseteq \mathbb{B}^{M_{t}*M_{t}}$ and $\theta_{t}\in\Theta_{t}\subseteq \mathbb{B}^{N*\left ( M_{t}+1 \right )}$:
\begin{align}
\gamma_{t}\in\Gamma_{t}  = &\left \{ s_{m}^{m^{'}} |s_{m}^{m^{'}}\in \left \{ 0,1 \right \} \quad m,m^{'}\in M^{*}_{t} \right \}
\label{eq:Gamma}
\end{align}
\begin{align}
\theta_{t}\in\Theta_{t}  = &\left \{ a_{n}^{m} |a_{n}^{m}\in \left \{ 0,1 \right \}\quad m\in M_{0,t}^{*},n\in N^{*} \right\}
\label{eq:Theata}
\end{align}
where $s_{m}^{m^{'}}$ suggests that, if the detected parts $m$ and $m^{'}$ can be used to form an individual mouse, \textit{i.e.,} $s_{m}^{m^{'}}=1$ with $m,m^{'}\in M^{*}_{t}$.

Our approach to the second problem is to assign correct identities to the detection candidates. In an easy case, the relationship between the targets and the detection candidates is bijective, which means that all the tracked targets are also observed and each measurement was generated by the tracked targets. This is an unrealistic scenario as it does not consider the false detection and the targets' occlusions. In order to solve this assignment problem, we introduce a placeholder for a `fake' (or missed) detection and also define $m \in M_{0}^{*}=\left \{0, 1,...,M \right \}$ to include the `fake' detection and all the detection candidates. Similar to $s_{m}^{m^{'}}$, a binary variable $a_{n}^{m}$ represents that the detection candidate index $m \in M^{*}_{0,t}$ which is generated by target $n \in N^{*}=\left \{ 1,...,N \right \}$ at time $t$. Here, we have $N*\left(M+1\right) $ paired with $a_{n}^{m}=1$ if candidate $m$ is assigned to target $n$ and $a_{n}^{m}=0$ otherwise. By definition, $\Gamma_{t}$ and $\Theta_{t}$ consist of all possible solutions at time $t$.

To ensure that every solution is physically available, we add several constraints that \textit{(1)} each candidate (except for the fake hypothesis $m=0$) is assigned to at most one target, i.e., (a) for $\forall m\in M^{*}_{t}, \sum_{n=1}^{N}a_{n}^{m}\leq 1$. \textit{(2)} each target is uniquely assigned to a candidate, i.e., (b) for $\forall n\in N^{*}_{t}, \sum_{m=0}^{M_{t}}a_{n}^{m}= 1$, and \textit{(3)} only the candidates and targets of the same type (e.g. head to head or tail to tail) can be corresponded, except for a fake hypothesis, i.e., (c) for $\forall m\in M^{*}_{t}, \forall n\in N^{*},a_{n}^{m}= 0$ if $\hat{m}\neq \hat{n}$.
In order to ensure that feasible solutions $\theta_{t}\in\Theta_{t}$ and $\gamma_{t}\in\Gamma_{t}$ result in a valid target assignment and pair-wise part association, we need to apply an additional constraint that two detected parts are connected if and only if both detections are assigned to different targets: (d) for $\forall m,m^{'}\in M^{*}_{t},s_{m}^{m^{'}}\leq\sum_{n=1}^{N}a_{n}^{m} \wedge s_{m}^{m^{'}}\leq\sum_{n=1}^{N}a_{n}^{m^{'}}$.

\subsection{Bayesian-inference Integer Linear Program}
We jointly resolve the targets assignment and part association problems by minimizing the following cost function:
\begin{equation}
\begin{split}
\mathcal{L}_{t}&=\min_{\theta{t} \in \Theta,\gamma_{t}\in\Gamma} -\left (\log \left (p\left (\theta_{t}\right )\right )+ \log \left (p\left (\gamma_{t})\right )\right )\right )
\end{split}
\label{eq:loss_function}
\end{equation}
where the values of $\theta_{t}$ and $\gamma_{t}$ which attain the minimum value of Eq. (\ref{eq:loss_function}) are the maximum likelihood of target assignment and part association at time $t$.

\subsubsection{Bayesian Inference}
With regards to the target tracking problem, the assumption of Markov property in the target state sequence is frequently employed in Extended Kalman filter \cite{mitzel2011real}, particle filtering \cite{khan2004mcmc,breitenstein2009robust}, MCFPHD \cite{wojke2016global} or MDP\cite{xiang2015learning}. These approaches are appropriate for the task of online tracking due to their concern on the region neighbourhood. We also utilize this assumption to resolve the target assignment problem. Let $V_{t}=\left \{ v^{1}_{t},v^{2}_{t},...,v^{N}_{t} \right \}$ denote the states of all the $N$ targets. State vector $v^{n}$ contains dynamic information of the position and velocity of the $j^{th}$ target at time $t$. $D_{t}=\left \{ d^{1}_{t},d^{2}_{t},...,d^{M}_{t} \right \}$ represent the positions of all the $M$ part detection candidates at time $t$. The assumption of Markov property is twofold. First, the kinematic state dynamics follow a first-order Markov chain: $p\left(v_{t}^n|v_{t-1}^n,...,v_{1}^n\right)=p\left(v_{t}^n|v_{t-1}^{n}\right)$, meaning that the state $v_{t}^n$ only depends on state $v_{t-1}^n$. Second, each observation $\hat{d}_{t}^{n}$ is only related to its state corresponding to this observation: $p\left(\hat{d}_{t}^{n}|\hat{d}_{1:t}^{n}V_{1:t}\right)=p\left(\hat{d}_{t}^{n}|v_{t}^n\right)$. Then, we have the following formulations:
\begin{equation}
\begin{split}
&p\left(v_{t-1}^{n}\right)\sim \mathcal{N}\left(\hat{v}_{t-1}^{n},\Sigma_{t-1}^{n}\right)\\&
p\left(v_{t}^n|v_{t-1}^{n}\right)\sim \mathcal{N}\left(A*v_{t-1}^{n},\Omega\right)\\&
p\left(\hat{d}_{t}^{n}|v_{t}^{n}\right)\sim \mathcal{N}\left(C*v_{t}^{n},\Upsilon \right)
\end{split}
\label{eq:trans_p}
\end{equation}
where, $A$ and $C$ are our state transition (or motion) and observation models respectively. $\hat{v}_{t-1}^{n}$ and $\Sigma_{t-1}^{n}$ are the mean value and the error covariance matrix at state $v_{t-1}^{n}$. $\mathcal{N}$ is the normal distribution. $\Omega$ and $\Upsilon$ represent the covariances of the process and observation noise with the mean value of zero. $p\left(\hat{d}_{t}^{n}\right)$ is calculated by marginalizing $p\left(\hat{d}_{t}^{n}|v_{t}^{n}\right)$ and $p\left(v_{t}^{n}|v_{t-1}^{n}\right)$:
\begin{equation}
\begin{split}
p\left(\hat{d}_{t}^{n}\right)=\int p\left(\hat{d}_{t}^{n}|v_{t}^{n}\right)\int p\left(v_{t}^{n}|v_{t-1}^{n}\right)*p\left(v_{t-1}^{n}\right)\mathrm{d}v_{t-1}^{n}\mathrm{d}v_{t}^{n}
\end{split}
\label{eq:marginalize_p}
\end{equation}
However, in a realistic case, Markovian assumption based methods carry the associated danger of drifting away from the correct target, as they treat the targets as conditionally independent of one another. This risk can be mitigated by optimizing data assignment and using prior knowledge, as shown in \cite{chenouard2013multiple,perera2006multi,giebel2004bayesian}. In our approach, when dealing with the data association problem, we use a Geometric Model as the prior knowledge. Given $K$ training samples centered at a single mouse, let $X_{k}= \left \{x_{k}^{1},x_{k}^{2},...,x_{k}^{n}\right \}$ (where $k = 1,...,K)$ denote the locations of the mouse parts in $k^{th}$ training sample. $x_{k}^{i}$ is the location of the $i^{th}$ part in the $k^{th}$ training sample. Let $X_{k}$ refer to a geometric template and suppose that the locations of the mouse parts in a feasible solution $\theta$ is generated by one of our geometric templates $X_{k}$. We then expand $p\left (\theta_{t}\right )$ as follows:
\begin{equation}
\begin{split}
p\left (\theta_{t}\right )=\sum_{k=1}^{K}p\left (\theta_{t}|X_{k}\right )p\left (X_{k}\right )
\end{split}
\label{eq:p_theta1}
\end{equation}
where our collections of $K$ training samples have been introduced into the calculation of $p\left (\theta_{t}\right )$ before it can be marginalized out.

By conditioning on the geometric template $X_{k}$, the locations of the mouse parts can be treated as conditionally independent from one another. As the association probability of the fake detection is not constrained by our geometric template, we rewrite the first term of Eq. (\ref{eq:p_theta1}) as follows:
\begin{equation}
\begin{split}
p\left (\theta_{t}|X_{k}\right )=\prod_{n\in N^{*}}p_{t}\left (a_{n}^{0}\right )*\prod_{m \in M^{*}_{t},n\in N^{*}}\frac{p_{t}\left (X_{k}|a_{n}^{m}\right )p_{t}\left (a_{n}^{m}\right )}{p_{t}\left (X_{k}\right )}
\end{split}
\label{eq:p_theta2}
\end{equation}
where
\begin{equation}
\begin{split}
p_{t}\left(a_{n}^{m}\right)\propto\left\{\begin{array}{lcl}
      \left(\left(1-\hat{p}\left(d_{t}^{m}\right)\right)\beta\right)^{a_{n}^{m}} &\text{if} & m=0\\
\left(\hat{p}\left(d_{t}^{m}\right)*p\left(\hat{d}_{t}^{n}=d_{t}^{m}\right)\right)^{a_{n}^{m}} &\text{otherwise} &
        \end{array}
\right.
\end{split}
\label{p_a}
\end{equation}
where $\hat{p}\left(d_{t}^{m}\right)$ is the detection probability of the $m^{th}$ part detection candidate at time $t$ and $p\left(\hat{d}_{t}^{n}\right)$ is a probability distribution obtained from our motion model, $p_{t}\left(a_{n}^{m}\right)$ is an assignment probability representing that the detection candidate index $m\in M^{*}_{t}$ is generated by target $n\in N^{*}$ at time $t$. If $m=0$, which means the target $n$ is missing, the $p_{t}\left(a_{n}^{m}\right)$ can be estimated using an empirical parameter $\beta$ of the false detection density.

Combining Eqs. (\ref{eq:trans_p}),(\ref{eq:marginalize_p}), (\ref{eq:p_theta1}) and (\ref{eq:p_theta2}) yields:
\begin{equation}
\begin{split}
p\left (\theta_{t}\right )=&\left(\sum_{k=1}^{K}\prod_{m \in M^{*},n\in N^{*}}p_{t}\left (X_{k}|a_{n}^{m}\right)\right)*\left(\prod_{n\in N^{*}}\left(\beta\right)^{a_{n}^{0}}\right.\\& \left. *\prod_{m \in M^{*},n\in N^{*}}\left(\hat{p}\left(d_{t}^{m}\right)*\mathcal{N}\left(d_{t}^{m};C*A*\hat{v}_{t-1}^{n},\right.\right.\right.\\&\left.\left.\left.C*\left(A*\Sigma_{t-1}^{n}*A^{T}+\Omega\right)*C^{T}+\Upsilon\right)\right)^{a_{n}^{m}}\right)
\end{split}
\label{eq:p_theta3}
\end{equation}
where, $p\left (X_{k}|a_{n}^{m}\right )$ represents that, if candidate $m$ is assigned to target $n$, i.e., the location of target $n$ is known, how well the part location in the $k^{th}$ geometric template fits the location of this target. Computing the sum as shown in Eq. (\ref{eq:p_theta3}) is challenging due to a large amount of training samples. However, we notice that, if $p\left (X_{k}|a_{n}^{m}=1\right )$ is very small, it will be unlikely to contribute much to the final outcome. Thus, we mainly consider geometric templates with large $p\left (X_{k}|a_{n}^{m}=1\right )$. To achieve this, we firstly find the $O$ nearest neighbours of the body in the training samples to match $\widetilde{b}_{m}\in B=\left \{ b^{1}_{t},b^{2}_{t},...,b^{Y}_{t} \right \}$ in the convolutional feature space, where $B$ is a set of body candidates with $Y$ elements which are generated by the proposed detector at time $t$ and $\widetilde{b}_{m}$ is the top-scoring body candidates which contain the candidate $m$. Then the probability distribution $p\left (X_{k}|a_{n}^{m}\right )$ of the $m^{th}$ part can be estimated by fitting a Gaussian model to the location of this part in the O body neighbours. Different from most top-down methods \cite{cootes1995active,twining2001robust,de2012computerized} that first track the mouse and then estimate the locations of mouse parts, our approach establishes geometric relationships between the mouse parts and the mouse body within our probabilistic model and tracks the mouse parts directly based on our probabilistic model, therefore our approach tolerates partial occlusion of the mouse body. The selection of the body targets can be formulated as follows:
\begin{equation}
\widetilde{b}_{m}=\left\{\begin{array}{lcl}
      \argmax_{b_{t}^{y}\in B}\sigma_{m}\left(b_{t}^{y}\right) &\text{if} & \max_{b_{t}^{y}\in B}\sigma_{m}\left(b_{t}^{y}\right)\\&&\neq 0\\
null &\text{otherwise} &
        \end{array}
\right.
\label{eq:b_m}
\end{equation}
where
\begin{equation}
\begin{split}
\sigma_{m}\left(b_{t}^{y}\right)=\left\{\begin{array}{lcl}
      \hat{p}\left(b_{t}^{y}\right) &\text{if} & \text{region of candidate m falls}\\ & & \text{outside region of } b_{t}^{y} \text{ by at}\\ & &\text{most } \upsilon \text{ pixels}\\
0 &\text{otherwise} &
        \end{array}
\right.
\end{split}
\label{eq:sigma_m}
\end{equation}
where, $\hat{p}\left(b_{t}^{y}\right)$ is the detection probability and $null$ is an empty value. In our experiments, we set $\upsilon=10$. With these in hand, we approximate the first term in Eq. (\ref{eq:p_theta3}) as:

\begin{equation}
\begin{split}
\sum_{k=1}^{K}\prod_{m \in M^{*},n\in N^{*}}p_{t}\left (X_{k}|a_{n}^{m}\right)\approx&\sum_{k\in O}\prod_{m \in M^{*},n\in N^{*}}p_{t}\left (X_{k}|a_{n}^{m}\right)
\end{split}
\label{eq:sigma_prod}
\end{equation}
where the sum is now only taken over those $k\in O$. The term $p_{t}\left (X_{k}|a_{n}^{m}\right)$ is a 2D Gaussian distribution centred at the location $x^{k}_{m,n}$ with a user-defined variance. According to the proof of Supplementary C, we obtain:

\begin{equation}
\begin{split}
\sum_{k\in O}\prod_{m \in M^{*},n\in N^{*}}p_{t}\left (X_{k}|a_{n}^{m}\right)=\prod_{m \in M^{*},n\in N^{*}}\sum_{k\in O}p_{t}\left (X_{k}|a_{n}^{m}\right)
\end{split}
\label{eq:sigma_prod2}
\end{equation}
If $O\neq\emptyset$, we alternatively fit a Gaussian model to the location of each part in the $O$ body neighbours and use this equation to estimate the sum as shown in Eq. (\ref{eq:sigma_prod2}), given the type (e.g. head or tail base) of the candidate $m$. Otherwise, we treat the sum in Eq. (\ref{eq:sigma_prod2}) as a very small variable value of $\varepsilon=0.0001$. We reformulate Eq. (\ref{eq:p_theta3}) as follows:
\begin{equation}
\begin{split}
p\left (\theta_{t}\right )\approx&\prod_{n\in N^{*}}\left(\beta\right)^{a_{n}^{0}}*\prod_{m \in M^{*},n\in N^{*}} \left(\Delta(a_{n}^{m})*\hat{p}\left(d_{t}^{m}\right)*\right.\\&\left.\mathcal{N}\left(d_{t}^{m};C*A*\hat{v}_{t-1}^{n},C*\left(A*\Sigma_{t-1}^{n}*A^{T}+\Omega\right)*C^{T}\right.\right.\\&\left.\left.+\Upsilon\right)\right)^{a_{n}^{m}}
\end{split}
\label{eq:p_theta4}
\end{equation}
where $\Delta(a_{n}^{m})$ defines a scoring function, estimated by our proposed geometric model given the location of the candidate $m$ in the body target.

\subsubsection{Parts association model}

To improve the multi-target tracking accuracy, several methods (\hspace{1sp}\cite{qin2012improving,yamaguchi2011you,pellegrini2009you}) have also presented group models, in which each object is considered as having the relationship with other objects and surroundings. These models can alleviate performance deterioration in crowded scenes. Similarly, we here propose a new part association model to establish the pair relationship between the detection candidates. In our approach, the pair-wise term $p\left (\gamma_{t}\right) $ shown in Eq. (\ref{eq:loss_function}) is defined as follows:
\begin{equation}
\begin{split}
p\left (\gamma_{t}\right )=\prod_{m ,m'\in M^{*}}\eta\left (s_{m}^{m'}\right )^{s_{m}^{m'}}
\end{split}
\label{eq:p_gamma}
\end{equation}
where, $\eta\left (s_{m}^{m'}\right)=p_{t}\left(d^{m}_{t},d^{m'}_{t}\right)/\left(1-p_{t}\left(d^{m}_{t},d^{m'}_{t}\right)\right)$. $p_{t}\left(d^{m}_{t},d^{m'}_{t}\right)\in \left(0,1\right)$ corresponds to the probability that detections $d^{m}_{t}$ and $d^{m'}_{t}$ in a frame t belong to the same mouse.  Note that $log\left(\eta_{t}\left (s_{m}^{m'}\right)\right)$ is negative if $p_{t}\left(d^{m}_{t},d^{m'}_{t}\right)>0.5$. Since we aim to minimize the cost function shown as Eq. (\ref{eq:loss_function}), a smaller  $log\left(\eta_{t}\left (s_{m}^{m'}\right)\right)$ i.e., a pair detection with a higher parts association probability are preferred. The parts association probability $p_{t}\left(d^{m}_{t},d^{m'}_{t}\right)$ depends on the types $\hat{m}$ and $\hat{m'}$ of the part detections $d^{m}_{t}$ and $d^{m'}_{t}$. If $\hat{m}=\hat{m'}$, we define $p_{t}\left(d^{m}_{t},d^{m'}_{t}\right)=IoU\left(d^{m}_{t},d^{m'}_{t}\right)$. This means that two close detections denoting the same part belong to the same mouse. If the connection between the two detections of the same type exists, the detections are merged with the weighted mean of the detections, where the weights are equal to $\hat{p}\left(d_{t}^{m}\right)$. If $\hat{m}\neq\hat{m'}$, we estimate the posterior probability $p_{t}\left(d^{m}_{t},d^{m'}_{t}\right) = p\left(s_{m}^{m'}=1|dis\left(d^{m}_{t},d^{m'}_{t}\right)\right)$ based on the Euclidean distance between the two detections. Assuming the prior probability $p\left(s_{m}^{m'}=1\right)=p\left(s_{m}^{m'}=0\right)=0.5$, then we have
\begin{equation}
\begin{split}
p&\left(s_{m}^{m'}=1|dis\left(d^{m}_{t},d^{m'}_{t}\right)\right)\\&= \frac{p\left(dis\left(d^{m}_{t},d^{m'}_{t}\right)|s_{m}^{m'}=1\right)}{p\left(dis\left(d^{m}_{t},d^{m'}_{t}\right)|s_{m}^{m'}=1\right)+p\left(dis\left(d^{m}_{t},d^{m'}_{t}\right)|s_{m}^{m'}=0\right)}
\end{split}
\end{equation}
where $p\left(dis\left(d^{m}_{t},d^{m'}_{t}\right)|s_{m}^{m'}=1\right)$ denotes the positive likelihood where two detections come from the same mouse. We estimate it by creating a 1-D Gaussian distribution of $dis\left(d^{m}_{t},d^{m'}_{t}\right)$ from the positive training examples. The negative likelihood $p\left(dis\left(d^{m}_{t},d^{m'}_{t}\right)|s_{m}^{m'}=0\right)$ is estimated in a similar way by using the negative training examples. With the above definition, $\hat{m}=\hat{m'}$ pairwise terms facilitate merging the mouse part candidates with the same type candidates. $\hat{m}\neq\hat{m'}$ pairwise terms allow us to connect the mouse part candidates within a valid mouse pose.
\subsubsection{Optimization}
By introducing Eqs. (\ref{eq:p_theta1}) and (\ref{eq:p_gamma}), we rewrite the cost function of Eq. (\ref{eq:loss_function}) as follows:
\begin{align}
\mathcal{L}&=\min -\left (
\sum_{n\in N^{*}}a_{n}^{0}*\log \beta+\sum_{m \in M^{*}_{t},n\in N^{*}} a_{n}^{m}*\log\left(\Delta(a_{n}^{m})*\nonumber\right.\right.\\&\left.\left.\hat{p}\left(d_{t}^{m}\right)*\mathcal{N}\left(d_{t}^{m};C*A*\hat{v}_{t-1}^{n},C*\left(A*\Sigma_{t-1}^{n}*A^{T}+\Omega\right)\nonumber \right.\right.\right.\\& \left.\left.\left.*C^{T}+\Upsilon\right)\right)+\sum_{m ,m'\in M^{*}_{t}}s_{m}^{m'}\log p_{t}\left (s_{m}^{m'}\right )\right )\nonumber\\&\text{s.t. \quad  (a),(b),(c),(d)\quad in section 4.1}
\label{eq:loss_function2}
\end{align}
Finally, the problems of targets assignment and part association are jointly reformulated as:
\begin{equation}
\begin{split}
\min_{\Lambda \in \left\{0,1\right\}^J} \Phi^T\Lambda \quad\quad s.t.\quad A y \leq h
\end{split}
\label{eq:min}
\end{equation}
where $\Lambda = \left[\theta,\gamma\right]^{T}$ is a binary vector of length $J=N\left(M+1\right)+M*M$, and $\Phi=\left[\Phi_{1},...,\Phi_{J}\right]^T$ is the cost vector with  $\Phi_{j}=-\log\left(\beta\right)$ if $1\leq j \leq M$, $\Phi_{j}=-\log\left(\Delta(a_{n}^{m})p_{t}\left(d_{t}^{m}\right)\mathcal{N}\right)$ if $M < j \leq N\left(M+1\right)$ and $\Phi_{j}=-\log\left(p_{t}\left(s_{m}^{m'}\right)\right)$ if $N\left(M+1\right)< j \leq N\left(M+1\right)+M*M$. $A$ and $h$ stand for a constraint matrix and linear equality constraints respectively, resulting in constraints (a), (b), (c) and (d).

To solve Eq. (\ref{eq:min}), we first relax the constraint $s_{m}^{m^{'}}\in \left \{ 0,1 \right \}, a_{n}^{m} |a_{n}^{m}\in \left \{ 0,1 \right \}$ to $0\leq s_{m}^{m^{'}} \leq 1, 0\leq a_{n}^{m} \leq 1$. Then, we utilize a branch-and-bound \cite{lawler1966branch} based global optimization method which is summarized and shown in Supplementary E. With the convergence analysis of the proposed algorithm shown in Supplementary E, it is proved that the algorithm converges to the global optimum.

The time complexity to solve Eq. (\ref{eq:min}) is $\mathcal{O}\left(\mathcal{A}^{1.5}\mathcal{B}^{2}\right)$ \cite{nemirovski2004interior}, where $\mathcal{B}=N*\left(M+1\right) + M*M$ denotes the number of the variables and $\mathcal{A}=N+M+1+M*M$ denotes the number of the constraints. 

\subsection{Targets state update}

After having obtained an optimal solution of Eq. (\ref{eq:min}), we associate each target with its corresponding detection candidate. If a target is assigned to a fake detection candidate, we assume that the target is occluded or lost. In order to guarantee tracking identification of each mouse part, we do not discard these targets like JPDAm \cite{hamid2015joint} and MDP \cite{xiang2015learning} (see their tracking results in Fig.S8 and S9). Instead we predict the target state using the motion model $\hat{v}_{t}^{n}=A*\hat{v}_{t-1}^{n}$ till the target is detected again. If a target is assigned to a true detection candidate, our goal is to maximize the posterior probability of each target state $\hat{v}_{t}^{n}$ given the detection $d_{t}^{m}$. The updated formula can be represented as follows:
\begin{equation}
\hat{v}_{t}^{n}=
\argmax_{v_{t}^{n}}p\left(v_{t}^{n}|d_{t}^{m}\right)
\label{eq:v_t}
\end{equation}
To compute the posterior probability $p\left(v_{t}^{n}|d_{t}^{m}\right)$, we firstly define
\begin{equation}
P=A*\Sigma_{t-1}^{n}*A^{T}+\Omega
\end{equation}
Following the Bayes' rule, we derive the posterior probability as follows:
\begin{equation}
\begin{split}
p\left(v_{t}^{n}|d_{t}^{m}\right)\sim &\mathcal{N}\left(\left(P^{-1}+C^{T}\Upsilon^{-1}C\right)^{-1}\left(C^{T}\Upsilon^{-1}d_{t}^{m}+\right.\right.\\&\left.\left.P^{-1}A\hat{v}_{t-1}^n\right),\left(P^{-1}+C^{T}\Upsilon^{-1}C\right)^{-1} \right)
\end{split}
\label{eq:p_post}
\end{equation}
where parameters $\mathcal{N}, C \text{ and } \Upsilon$ have been defined as those of Eq. (\ref{eq:trans_p}).

Afterwards, the estimated state $\hat{v}_{t}^{n}$ is the mean of the Gaussian distribution:
\begin{equation}
\begin{split}
\hat{v}_{t}^{n}=
\left(P^{-1}+C^{T}\Upsilon^{-1}C\right)^{-1}\left(C^{T}\Upsilon^{-1}d_{t}^{m}+P^{-1}A\hat{v}_{t-1}^n\right)
\end{split}
\label{eq:v_t2}
\end{equation}
Note that we solve the part tracking problem of multi-mice in an online manner, which is illustrated in Supplementary F and is summarized in Alg. \ref{Alg:tracking}.

\begin{algorithm}[H]
\caption{Algorithm for online mouse part tracking.}
\begin{algorithmic}[1]
\renewcommand{\algorithmicrequire}{\textbf{Input:}}
\renewcommand{\algorithmicensure}{\textbf{Output:}}
\REQUIRE a video sequence $\mathcal{F}$, the proposed mouse part and body detector, $C$ training samples of a single mouse.
\ENSURE bounding boxes of all the targets.
\FOR {$t = 1$ to $T$}
 \STATE Generate a set of part and body detection candidates using the proposed mouse part and body detector on the frame $\mathcal{F}_{t}$
 \STATE Fit a geometric model with appearance features to $O$ neighbours of the top-scoring body candidate, which contain candidate $m$;
 \STATE Compute $\Delta \left(a_{n}^{m}\right)$, $\mathcal{N}\left(d_{t}^{m}\right)$ and $p_{t}\left(s_{m}^{m'}\right)$ in Eq. (\ref{eq:p_theta4});
 \STATE Construct constraint matrix $A$ in Eq. (\ref{eq:min}) using the constraint formula (a), (b), (c) and (d);
 \STATE Construct cost vector $\Psi$ in Eq. (\ref{eq:min});
\STATE Obtain best solution of $\theta$ and $\gamma$ by solving Eq. (\ref{eq:min}) in Alg. \ref{Alg:optimization};
\STATE Update the state of each target using Eq. (\ref{eq:v_t2});
\STATE Obtain bounding boxes of all the targets dependent on the best solution of $\theta$ and $\gamma$;
\ENDFOR
\RETURN bounding boxes of all the targets.
\end{algorithmic}
\label{Alg:tracking}
\end{algorithm}

\section{Experimental Set-up}
\subsection{Our Multi-Mice Parts Track Dataset}
In this paper, we introduce our new dataset for multi-mice part tracking in videos. The dataset was collected in collaboration with biologists of Queen's University Belfast, United Kingdom, for a study of neurophysiological mechanisms involved in Parkinson's disease. In our dataset, two or three mice are interacting freely in a 50*110*30cm home cage and are recorded from the top view using a Sony Action camera (HDR-AS15) with a frame rate of 30 fps and 640 by 480 pixels VGA video resolution. All the experiments are conducted in an environment-controlled room with constant temperature (27$^{\circ}$C) and light condition (long fluorescent lamp 40W). The dataset provides the detailed annotations for multiple mice in each video, as shown in Supplementary G. The mice used throughout this study were housed under constant climatic conditions with free access to food and water. All the experimental procedures were performed in accordance with the Guidance on the Operation of the Animals (Scientific Procedures) Act, 1986 (UK) and approved by the Queen’s University Belfast Animal Welfare and Ethical Review Body. Our database covers a wide range of activities like contacting, following and crossing. Moreover, our database contains a large amount of mouse pose and mouse part occlusion. After proper training, six professionals were invited to annotate mouse heads, tail bases and localise each mouse body in the videos. We assign a unique identity to every mouse part appearing in the images. If a mouse part was in the field-of-view but became invisible due to occlusion, it is marked `occluded'. Those mouse parts outside the image border limits are not annotated. In total, our dataset yields 5 videos of two mice and 5 videos of three mice, each video lasts 3 minutes and 400 frames randomly sampled from each video are annotated (4000 frames are annotated in total). Since the problem of multi-mice parts has not been quantitatively evaluated in the literature, we follow the evaluation metrics defined in multi-target tracking \cite{everingham2010pascal,ristani2016performance} for this problem, and compare the results with several baseline methods. We split the dataset into training and testing sets with an equal duration of time and train our network based on transfer learning with pre-trained models. The code and the dataset are published on github: \url{https://github.com/ZhehengJiang/BILP}.

\subsection{Evaluation metrics}
In order to evaluate the proposed mouse part detector, we use the widely adopted precision, recall, average precision (AP) and mean average precision (mAP)\cite{everingham2010pascal}. For a specific class, the precision value corresponds to the ratio of the positive object detections against the total number of the objects that the classifier predicts, while the recall value is defined as the percentage of the positive object detections against the total number of the objects labelled as ground-truth. The precision-recall curves are obtained by varying the model score threshold in the range of 0 and 1, which determines what is counted as a positive detection of the class. Note that, with our metrics, only the object detections that have an IoU ratio over 0.5 with the ground-truth are counted as positive detections while the rest are negatives. The AP score is defined as the average of precision at the set of 11 equally spaced recall values. mAP is just the average over all the classes. To provide a fair comparison for both types of occlusion handling, we consider an occluded mouse part correctly estimated either if (a) it is predicted at the correct location despite being occluded, or (b) it is not predicted at all. Otherwise, the prediction is considered as a false positive.

To evaluate the part tracking performance, we consider each mouse part trajectory as one individual target, and compute the multiple object tracking precision (MOTP) and the multiple object tracking accuracy (MOTA)\cite{bernardin2008evaluating}. The former is derived from three types of error ratios: false positives (FP), missed targets (MT), and identity switches (IDs). These errors are normalized by the number of the objects appearing in the image frames and can be summed up to produce the resulting tracking accuracy, where 100\% corresponds to zero errors. MOTP measures how precise each mouse part has been localized. We also report the trajectory-based measures of the number of mostly tracked (MT) and mostly lost (ML) targets. If a track hypothesis has covered at least 80\% of its life span based on the ground truth trajectories, it is considered as MT. If less than 20\% are not tracked, the track hypothesis is considered as ML. IDF1 \cite{ristani2016performance} measures the ratio of the correctly identified detections. OSPA computes the optimal subpattern assignment metric between a set of tracks and the ground truth.
\section{Experimental Results}
In this section, we evaluate the proposed method for part tracking on the newly introduced Multi-Mice PartsTrack dataset.

\subsection{Evaluation of Part and Body Detector}
\subsubsection{Implementation}
The multiple scales and aspect ratios of the root boxes in our method are different for mouse parts and body. Root boxes of inappropriate scales and aspect ratios are ineffective for mouse detection. For mouse `head' and `tail base', we choose a single aspect ratio of 1.13 (width to height) and four scales with the root box widths of 24, 29, 35 and 42 pixels based on the statistics of the object shape in the training set. For mouse `body' cases, we use multiple aspect ratios of 0.5 (landscape),1 (square) and 2 (portrait), and three scales with the root box widths of 50, 80 and 128 pixels. We label a root box as a positive example if its IoU ratio is greater than 0.7 with one ground-truthed box, and a negative example if its IoU ratio is lower than 0.3 with all the ground-truthed boxes. Root boxes that are neither positive nor negative do not contribute to the training RPNs. This experiment is conducted on the 2-mice dataset.
\begin{table}
\begin{center}
\caption{Performance (precision) of the proposed part detector using different methods.}
\label{tab:pre-trained_models}
\begin{tabular}{p{1.3cm}p{1.2cm}p{1.1cm}p{0.5cm}p{0.5cm}p{0.5cm}p{0.5cm}}
\toprule
method&Pre-trained Model&Layers & head & tail base & body & mAP \\
\hline
\multirow{2}{1.2cm}{faster rcnn~\cite{ren2015faster}}& \textit{ResNet50}&/& 93.5\% & 76.0\% & 97.1\% & 88.9\% \\
&\textit{ResNet101}&/&96.3\% &80.2\% &95.6\% &90.7\% \\
\hline
\textit{ssd300}~\cite{liu2016ssd}& \textit{VGGNet16}\cite{simonyan2014very}&/&86.9\% &63.7\%&98.1\%&82.9\%\\
\textit{ssd512}~\cite{liu2016ssd}& \textit{VGGNet16}\cite{simonyan2014very}&/&90.6\% &75.4\%&98.2\%&88.1\%\\
\hline
\textit{YOLO}\cite{redmon2016you}& \textit{Darknet}\cite{redmon2016you}&/&93.6\%&77.3\%&98.6\%&89.8\%\\
\hline
\textit{Motr}\cite{ohayon2013automated}& /&/&/&/&82.4\%&/\\
\hline
Blob detection\cite{giancardo2013automatic}& /&/&/&/&89.8\%&/\\
\hline
\multirow{10}*{\textit{ours}}&\multirow{3}{1.4cm}{\textit{AlexNet}~\cite{krizhevsky2012imagenet}}& 5 layers & 88.8\% & 44.8\% & 86.9\% & 73.5\%\\
&&9 layers & 49.5\% & 25.7\% & 89.5\% & 54.9\%\\
&&15 layers & 35.2\% & 12.4\% & 86.8\% & 44.8\%\\
\cline{2-7}
&\multirow{3}{1.4cm}{\textit{VGGNet16}\\~\cite{simonyan2014very}}& 6 layers & 89.9\% & 67.5\% & 77.3\% & 78.2\%\\
&&11 layers & 91.5\% & 74.2\% & 90.3\% & 85.3\%\\
&&18 layers & 85.8\% & 51.0\% & 89.8\% & 75.5\%\\
\cline{2-7}
&\multirow{3}{1.4cm}{\textit{VGGNet19}\\~\cite{simonyan2014very}}&6 layers & 91.5\% & 64.0\% & 87.3\% & 80.9\%\\
&&11 layers & 95.6\% & 78.4\% & 95.4\% & 89.8\%\\
&&15 layers & \textbf{98.4}\% & \textbf{89.4}\% & \textbf{98.3\%} & \textbf{95.4\%}\\
&&20 layers & 76.8\% & 42.1\% & 91.5\% & 70.1\%\\
\bottomrule
\end{tabular}
\end{center}
\end{table}

\subsubsection{Results}
We firstly investigate the performance of the proposed part and body detector using different pre-trained models. All the experiments adopt a 3-stage Part Proposal Network based on the trade-off between speeds and system performance, as shown in Supplementary H Fig.S3. From Table \ref{tab:pre-trained_models}, we observe that deeper network (VGG19) has achieved superior performance over other shallow networks (i.e., AlexNet and VGG16). It is because the deeper network can learn richer image representations. But surprisingly, the accuracy of the mouse parts is degraded after using more than 9 layers of AlexNet, 11 layers of VGGNet16 or 15 layers of VGGNet19. This limitation is partially because of the low-resolution features in the feature map of the higher layer. These features are not discriminative on the small regions, and thus degrade the performance of the downstream classifier. In comparison, these features are cooperative enough to distinguish the mouse body from the background as the region area of the mouse body is 3 times larger than that of the mouse head and the tail base. This result also suggests that, if reliable features can be extracted, the downstream classifier is able to improve the detection accuracy. In addition, using the network with layers less than 6 layers of VGGNet 16 or 6 layers of VGGNet 19 starts to demonstrate accuracy degradation, which can be explained by the weaker representation ability of the shallower layers. In Table \ref{tab:pre-trained_models}, We have also compared DNN based detectors with two conventional methods\cite{ohayon2013automated,giancardo2013automatic},  which only provide methods to detect mouse body. The DNN based detectors show better performance than the conventional method in body detection.

\begin{table}
\begin{center}
\caption{Comparisons of different classifiers and features.}
\label{tab:different_features}
\begin{tabular}{p{1.5cm}p{1.5cm}p{1.5cm}p{1.2cm}p{1.2cm}}
\toprule
ROI Feature & Classifier & head & tail base & background \\
\hline
\multirow{2}*{LBP}  & linear SVM & 93.1\% & 92.2\% & 94.7\%\\
 & fc layers & 95.8\% & 94.3\% & 95.1\%\\
\multirow{2}*{HOG}  & linear SVM & 93.2\% & 85.3\%& 83.4\%\\
 & fc layers & 94.1\% & 87.2\% & 83.7\%\\
\multirow{2}*{SIFT + FV} & linear SVM& 77.6\% & 72.4\% & 60.8\%\\
 & fc layers & 78.1\% & 73.8\% & 61.7\%\\
\multirow{2}*{Conv} & linear SVM  & 92.9\% & 91.1\%& 94.7\%\\
 & fc layers & 93.2\% & 92.3\% & 96.3\%\\
\multirow{2}*{Conv + LBP}& linear SVM & 96.5\% & 95.4\%& 95.3\% \\
 & fc layers & \textbf{98.8\%} & \textbf{97.3\%}& \textbf{96.0\%}\\
\multirow{2}*{Conv + HOG}& linear SVM & 93.3\% & 89.5\%&90.3\%\\
 & fc layers & 94.5\% & 90.6\%&92.3\%\\
\bottomrule
\end{tabular}
\end{center}
\end{table}

\begin{figure}
\begin{center}
\begin{tabular}{cc}
\includegraphics[width=4.4cm]{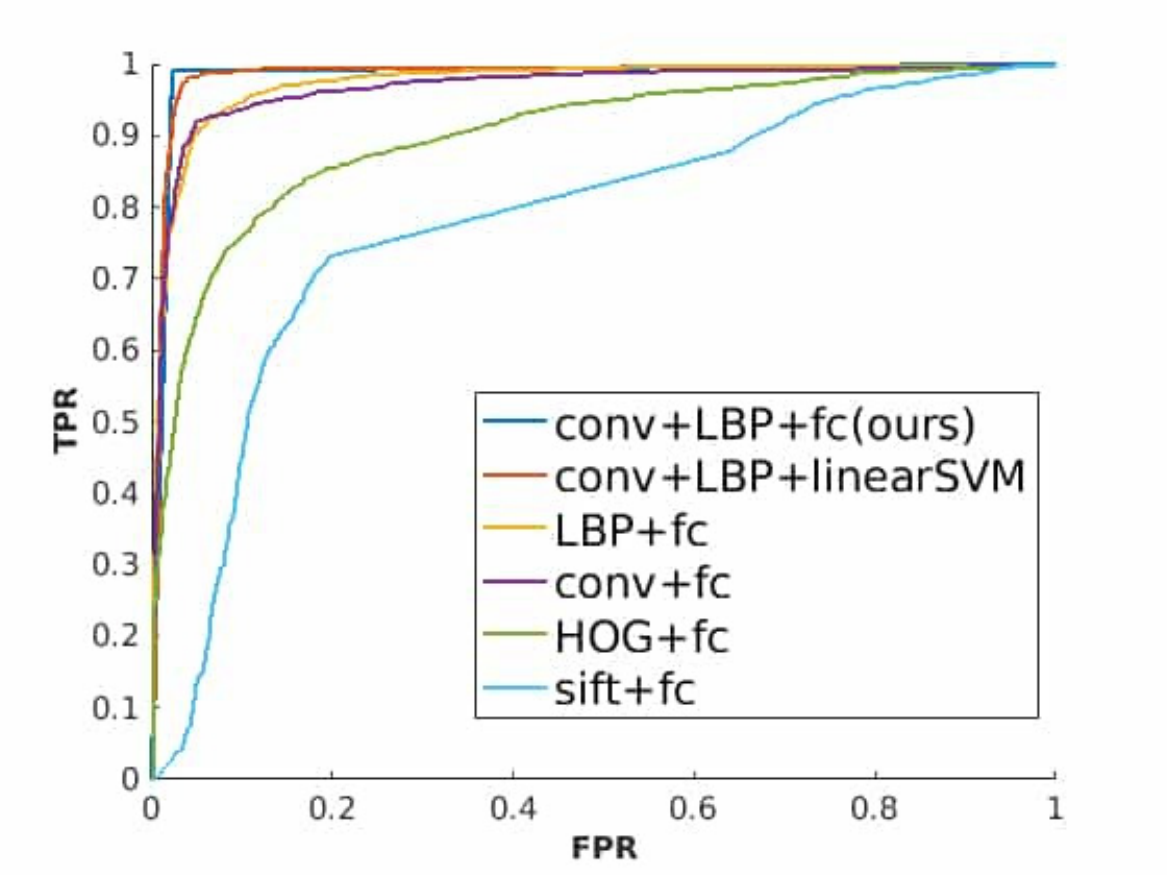} & \includegraphics[width=4.4cm]{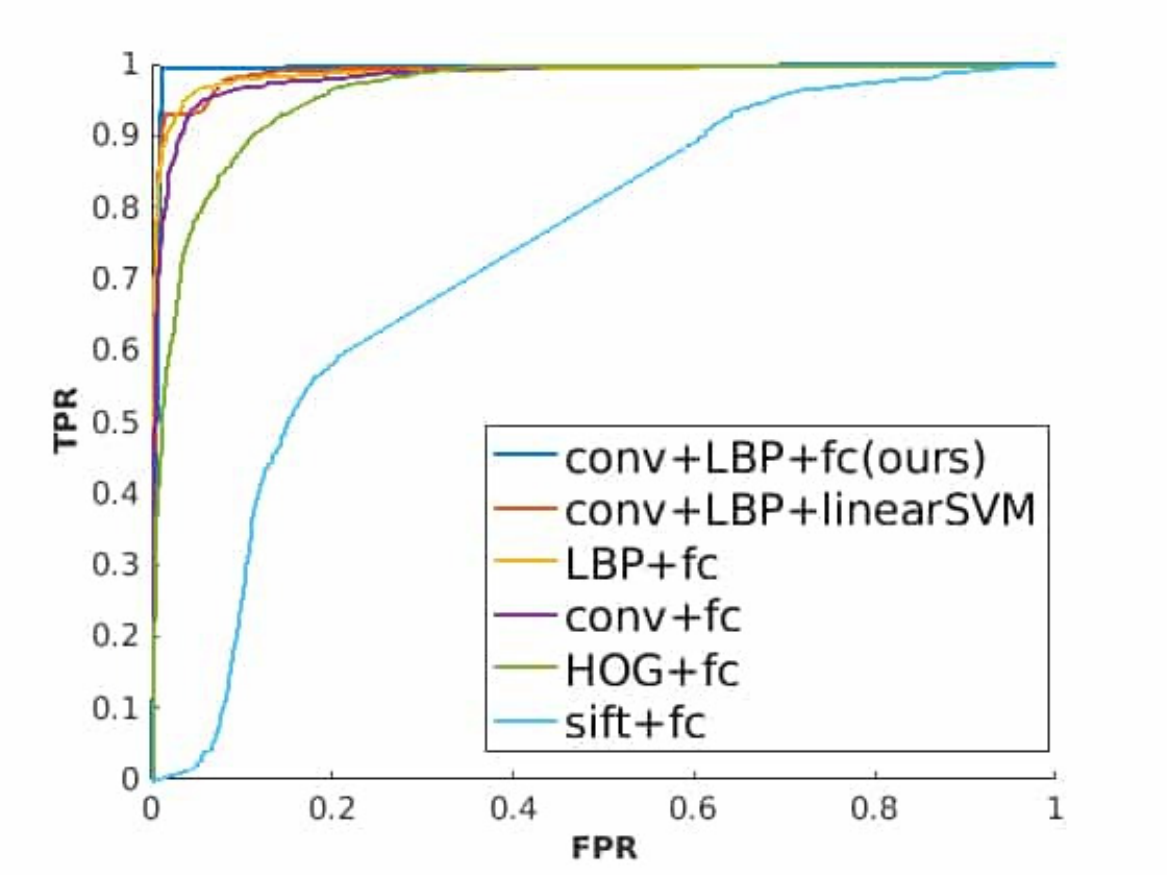}\\(a) head &
(b) tail
\end{tabular}
\end{center}
\caption{ROC curve shows the true positive rate (TPR) against the false positive rate (FPR) for different mouse parts.}
\label{fig:tail_head_roc}
\end{figure}

\begin{figure*}
\begin{center}
\begin{tabular}{ccc}
\includegraphics[width=5.8cm]{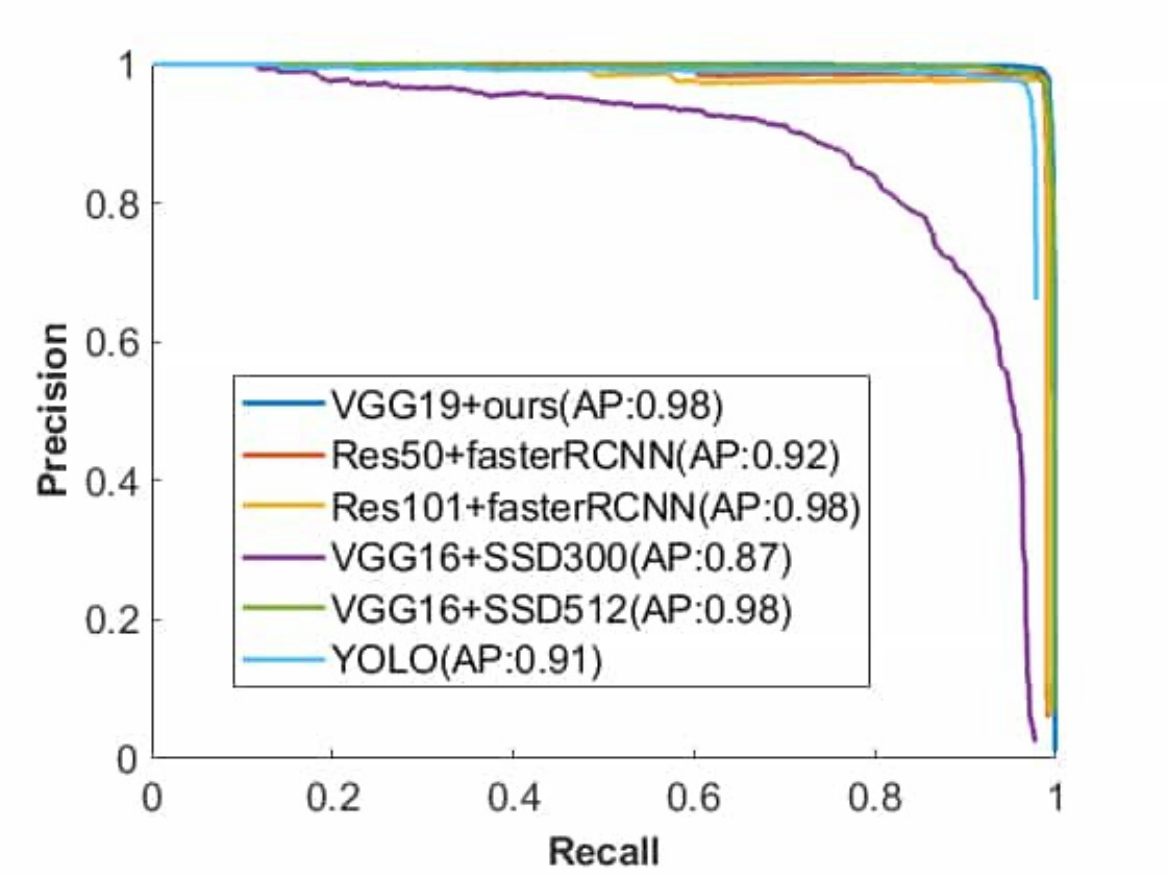}&
\includegraphics[width=5.8cm]{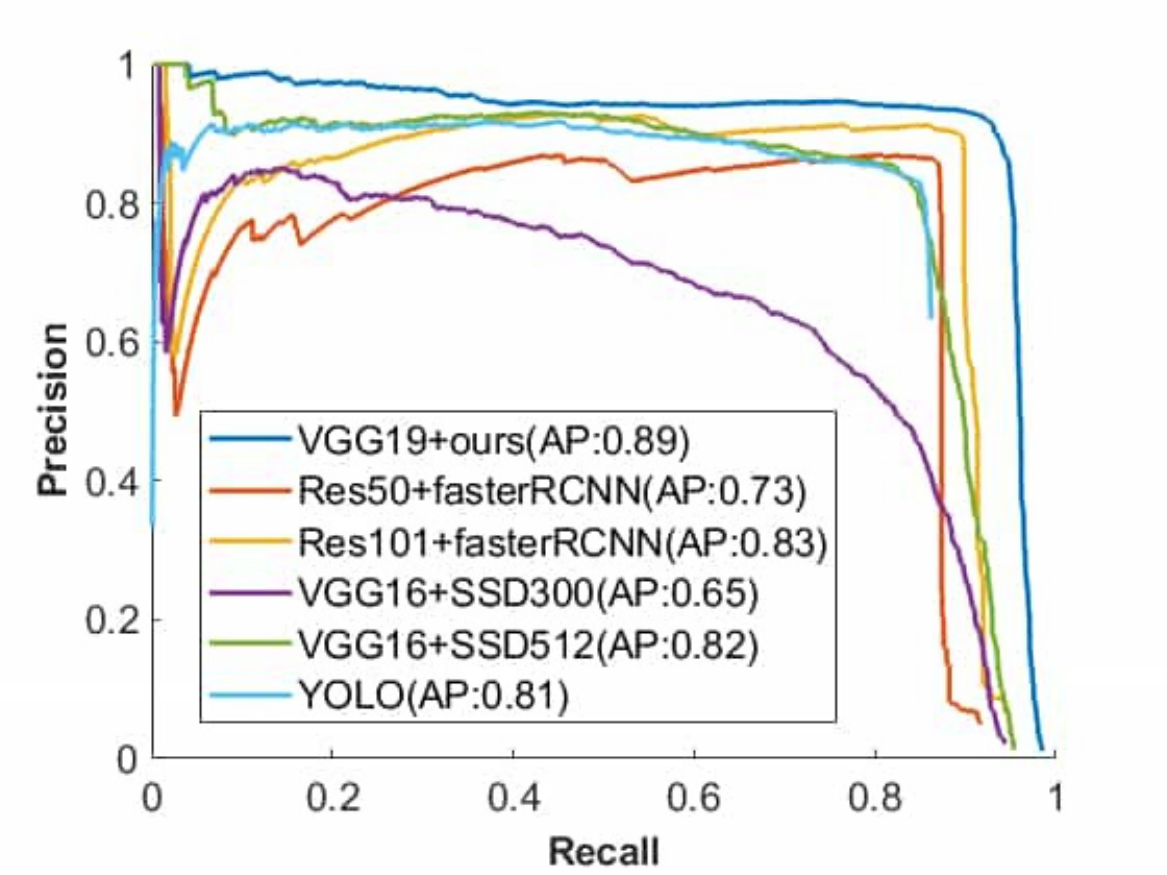}&
\includegraphics[width=5.8cm]{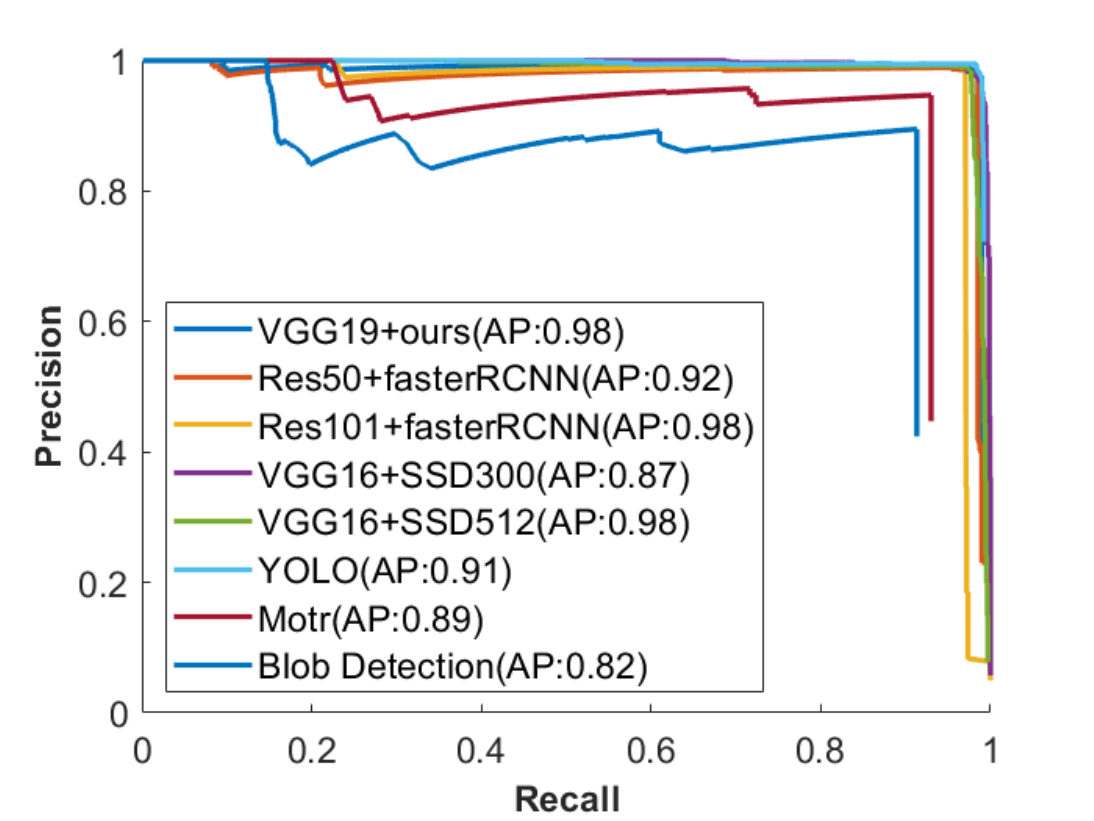} \\
(a) head&(b) tail&(c) body
\end{tabular}
\end{center}
\caption{Precision/Recall curves of different methods on the 2-mice dataset.}
\label{fig:diff_methods}
\end{figure*}

\begin{table*}
\begin{center}
\caption{Quantitative evaluation of multi-mice parts tracking on the 3-Mice dataset.}
\label{tab:tracking_results_3m}
\begin{tabular}{p{5.5cm}p{1cm}p{1cm}p{1cm}p{1cm}p{1cm}p{1cm}p{1cm}p{1cm}p{1cm}}
\toprule
Method\quad(speed) & MOTA & MOTP &  MT & ML & FP & FN & IDs & IDF1& OSPA\\
& $\%\uparrow$ & $\%\uparrow$ & $\uparrow$ &$\downarrow$&$\downarrow$&$\downarrow$&$\downarrow$&$\%\uparrow$&$\downarrow$\\
\midrule
\multicolumn{8}{c}{\textit{Impact of the constraints}}\\
\hline
All\quad(30.8 fps)  & \textbf{81.1} & 65.5  & \textbf{25} & \textbf{0} & \textbf{566} & \textbf{546} & \textbf{12} & \textbf{91.2}&\textbf{6.55}\\
All(b,c,d)  & 80.2 & 65.6  & 24 & 0 & 581 & 571 & 26 & 89.7&/\\
All(a,c,d)  & 79.5 & 65.6  & 25 & 0 & 615 & 595 & 12 & 90.75&/\\
All(a,b,d)  & 75.4 & 66.0 & 22  & 0 & 738 & 718 & 12 &  90.2 &/\\
All(a,b,c)  & 74.8 & 65.8  & 24 &  0 & 764 & 735 & 12 & 89.2&/\\
\hline
\multicolumn{8}{c}{\textit{Impact of the geometric model}}\\
\hline
All without geometric model \quad(34.7 fps) & 70.3 & 66.6  & 21 & 0 & 892 & 872 & 12 & 88.8&/\\
\hline
\multicolumn{8}{c}{\textit{Impact of the motion model}}\\
\hline
All without motion model \quad(31.6 fps) & 61.2 & 65.5  & 22 & 2 & 1101 & 1061 & 152 & 76.14&/\\
\hline
\multicolumn{8}{c}{\textit{Impact of the parts association model}}\\
\hline
All without parts association model \quad(31.2 fps) & 76.3 & 65.4  & 24 & 1 & 688 & 693 & 30 & 88.5&/\\
\hline
\multicolumn{8}{c}{\textit{Comparison with the state-of-the-art}}\\
\hline
MOTDT~\cite{long2018real}\quad(20.6 fps) & 20.5 & 31.4  & 21 & 0 & 1592 & 1152 & 3103 & 20.2&26.47\\
MDP~\cite{xiang2015learning}\quad(2.5 fps) & 64.6 & 70.2  & 21 & 0 & 1114 & 994 & 18 & /&21.26\\
SORT~\cite{bewley2016simple}\quad(260 fps) & 59.9 & 70.5  & 19 & 1 & 971 & 1331 & 106 & 15.8 & 15.42\\
JPDAm~\cite{hamid2015joint}\quad(32.6 fps) & 67.0 & 65.1  & 26 & 0 & 1264 & 548 & 155 & 44.6&20.14\\
RNN-LSTM~\cite{milan2017online}\quad(50 fps) & 50.6 & 67.3  & 20 & 0 & 1688 & 1176 & 102 & 54.6& 22.16\\
CEM~\cite{milan2013continuous}\quad(7.7 fps) & 34.1 & 69.5  & 21 & 1 & 2637 & 1093 & 224 & / &24.75\\
DeepSORT~\cite{wojke2017simple}\quad(20 fps) & 42.7 & 32.4  & 19 & 0 & 1796 & 1419 & 212 & 44.4 & 18.63\\
\bottomrule
\end{tabular}
\end{center}
\end{table*}

In Table \ref{tab:different_features}, we report the capability of different features to distinguish mouse parts and the background. Since CNN features perform better than the other hand-crafted features for body detection, we do not exploit any hand-crafted feature to maintain the system accuracy. For fair comparisons, all the proposals are generated from the same PPN which is initialized by the first 15 layers of VGG-19. We produce part proposals from images and extract low-level features such as LBP, HOG and SIFT. Interestingly, training our fc layers with LBP features on the same set of the PPN proposals actually has led to better average accuracy 95.1\%(vs. Conv's 93.9\% and HoG's 88.3\%), shown in Table \ref{tab:different_features}. Combining LBP and Conv features, we can achieve the best result 97.37\%. Similar results are also obtained using a linear SVM classifier. But the fc layers are more flexible and can be trained with other layers in an end-to-end style. Since too few interest points are detected from the mouse head and the tail base, SIFT features encoded by Fisher Vector~\cite{sanchez2013image} present a much worse result on this task. Figure \ref{fig:tail_head_roc} clearly shows
that our combined LBP and conv features lead to the best performance for classifying mouse 'head' and 'tail base' proposals.

We also compare our detection network against Faster R-CNN \cite{ren2015faster}, SDD~\cite{liu2016ssd} and YOLO \cite{redmon2016you} on our dataset. These methods are also trained based on transfer learning with pre-trained models. We train Faster R-CNN based on the two pre-trained models of Residual Network 50 and 101 \cite{he2016deep}. SSD~\cite{liu2016ssd} has two input sizes (300 $\times$ 300 vs. 512 $\times$ 512). The experiments show that the model with a larger input size leads to better results. YOLO uses its own pre-trained model named Darknet. As shown in Figure \ref{fig:diff_methods}, most methods have high APs on the detection of head and body. However, the performance of Faster RCNN, SSD and YOLO falls considerably when they are applied to detecting the tail base. Our detection method has better performance to detect the tail base and its AP degrades less than the other methods. We also observe that if the threshold of detection scores was previously set high, lowering the threshold in Faster RCNN, SSD and YOLO will introduce more false positives than our methods, resulting in a sharp decrease of precision as shown in Figure \ref{fig:diff_methods}.

\subsection{Evaluation of Multi-mice parts tracking}
\subsubsection{Implementation}
In our tracking algorithm, each part target's state contains position, velocity and shape of bounding box $x_{t}=(x_{p},x_{v},y_{p},y_{v},w_{p},h_{p})$. $(x_{p},y_{p})$, $(x_{v},y_{v})$ and $(w_{p},h_{p})$ respectively denote the position, the velocity and the shape of each part target. For the shape of each target, we use the width and height of the detected bounding box. We model the motion of each part target based on the linear dynamical system discretized in the time domain and predict the state of each part target in the next image frame as $x_{t}=A x_{t-1}+\mu$, where $A=diag(A_{x},A_{y})$ is a constant transition model, and $A_{x}=A_{y}=[1, \tau ;0, 1]$. $\mu$ is a system noise and subject to a Gausssian distribution with covariance $\Omega=diag(\Omega_{x},\Omega_{y})$, $\Omega_{x}=\Omega{y}=q_{d}[{\tau}^{3}/3, {\tau}^{2}/2;{\tau}^{2}/2,\tau]$. Here $\tau=1$ and $q_{d}=0.5$ refer to the sampling period and  the process noise parameter respectively. The uniform clutter density $\beta$ is estimated as $\beta=f_{false}/\left( w_{im}*h_{im}\right)$,$w_{im}=480$ and $h_{im}=640$ are the width and height of the image, $f_{false}=0.1$ is the average number of the false detections per image frame. This experiment is conducted on both our 2-Mice and 3-Mice datasets. We compare the proposed approach against six baselines following their default settings.

\subsubsection{Results}
The results of multi-mice part tracking are reported in Table \ref{tab:tracking_results_3m} and Supplementary I Table S2. We quantitatively  evaluate the proposed system using the commonly seen multi-object tracking metrics. The Up and Down arrows in the table indicate whether higher or lower values are obtained. To evaluate the proposed optimization objective function (Eq. (\ref{eq:loss_function})) for multi-mice part tracking, we have quantified the  impact of different constraints (a)-(d) during the optimization. To this end, we optimize the problem by removing one constraint at a time. As shown in Table \ref{tab:tracking_results_3m}, all the types of the constraints make evident contributions to the system performance, as they ensure that every solution is physically feasible. We also examine the impact of the geometric, motion and part association models. Removing one of the three components significantly decreases the performance. In particular, the motion model plays the most crucial role, which is obvious on the 3-mice Dataset. The part association and geometric models can effectively reduce ID switch and FP respectively. This is expected because these two models help to obtain the best target assignment and part association. We observe the change of our approach's speeds by removing different components and the geometric model has the biggest impact on speeds. We also implement several standard multi-target tracking methods on our multi-mice PartsTrack dataset. For fair comparison, we use the proposed PPN detector to generate part bounding boxes, and perform part tracking using several state-of-the-art part trackers \cite{bewley2016simple,hamid2015joint,xiang2015learning,long2018real,milan2017online,milan2013continuous}. As we can see, at the bottom of Table \ref{tab:tracking_results_3m} and Supplementary I Table S2, our tracker achieves the highest MOTA scores. Our method results in the lowest number of ID switches and highest IDF1 scores. This is primarily due to our powerful geometric and pair based part association models, which can handle part identities more robustly. We conduct ablation studies on the multi-stage training design shown in Fig. S3, showing the effectiveness of our training strategy. The established SORT \cite{bewley2016simple} has achieved the best efficiency due to the use of a simple target assignment model, where the cost function is fully dependent on IOU distance between each detection and the predicted bounding box. Evidence shows that this method cannot maintain consistent tracking performance (see Table \ref{tab:tracking_results_3m} and \ref{tab:tracking_results_2m}). Note that the speed of all approaches reported in Table \ref{tab:tracking_results_3m} and \ref{tab:tracking_results_2m} excludes the time of the part candidate detection. Fig. S5 and Tab. S2 show the tracking results of conventional methods (Motr and MiceProfiler), which track mice by fitting ellipses and hard-coded geometric models respectively. From Fig. S5, we observe that if the mice are very close, the models used in these methods cannot be properly fitted, leading to tracking targets’ swaps or incorrect part localization. Moreover, these models are hard-coded in the system, and thus limit the flexibility of their methods. Figs. S6, S8 and S9 in Supplementary F provide the qualitative comparison of the proposed tracking method against MDP \cite{xiang2015learning} and JPDAm \cite{hamid2015joint} using the same detection results. As shown in Fig. S6, MDP swap identities between targets 2 and 5 after the occlusion is caused by `Pinning' in the image of the middle column, while JPDAm assigns a new identity to target 6 after it is occluded by target 3. Similar problems can also be witnessed in Figs. S8 and S9. Compared to these two tracking methods, the proposed approach correctly integrates the detection results with the tacking practice and predicts the occlusion.

Overall, our approach is not the fastest in multi-mice tracking, but sustains the best accuracy. Fig. \ref{fig:tracking_results} shows exemplar tracking results of the test sequences in the proposed Multi-Mice PartsTrack dataset.

\subsection{Generalization to Other Datasets}
To demonstrate the generalization of our proposed approach, we also implement the approach on two public datasets related to other insects or animals: locusts were recorded in a laboratory setting and zebras recorded in the wild \cite{graving2019deepposekit}. The publicly available datasets of multiple animals recorded from the top view are scarce and the two datasets appearing here are examples. They contain 700 and 900 annotated frames for training and validation, 1300 and 1600 unlabeled frames for testing, respectively. Following \cite{graving2019deepposekit}, we split the annotated datasets into 90\% training examples and 10\% validation examples. Our detection approach achieves 100\% accuracy of detecting animal's heads, tails and bodies on both datasets. Since their testing datasets are unlabelled, we illustrate the performance of the proposed approach against these datasets in the supplementary videos. In comparison, our dataset seems more challenging, where the mouse parts are deformable with more occlusions during moving. Moreover, our annotated dataset is larger with a quantitative test-bed for part detection and tracking of multiple mice.


\section{Conclusion}
In this paper, we have presented a novel method for markerless multi-mice part detection and tracking. We have demonstrated that the proposed multi-stage part and body detector performed effective hard negative mining and achieved satisfactory detection results. We also proposed a new formulation based on target assignment with the learned geometric constraints and a pair-wise association scheme with motion consistency and restriction. Moreover, we presented a challenging annotated dataset to evaluate the proposed algorithms for multi-mice part tracking. Experimental results on the proposed datasets demonstrate that the proposed algorithm outperformed other state-of-the-art and conventional methods. We also demonstrate the generalization ability of our approach on other species. Our future work will address social behaviour analysis using the proposed tracking method.


\ifCLASSOPTIONcaptionsoff
  \newpage
\fi

\bibliographystyle{IEEEtran}
\bibliography{mybibfile}
\vspace{-8 mm}

\begin{IEEEbiography}[{\includegraphics[width=1in,height=1.25in,clip,keepaspectratio]{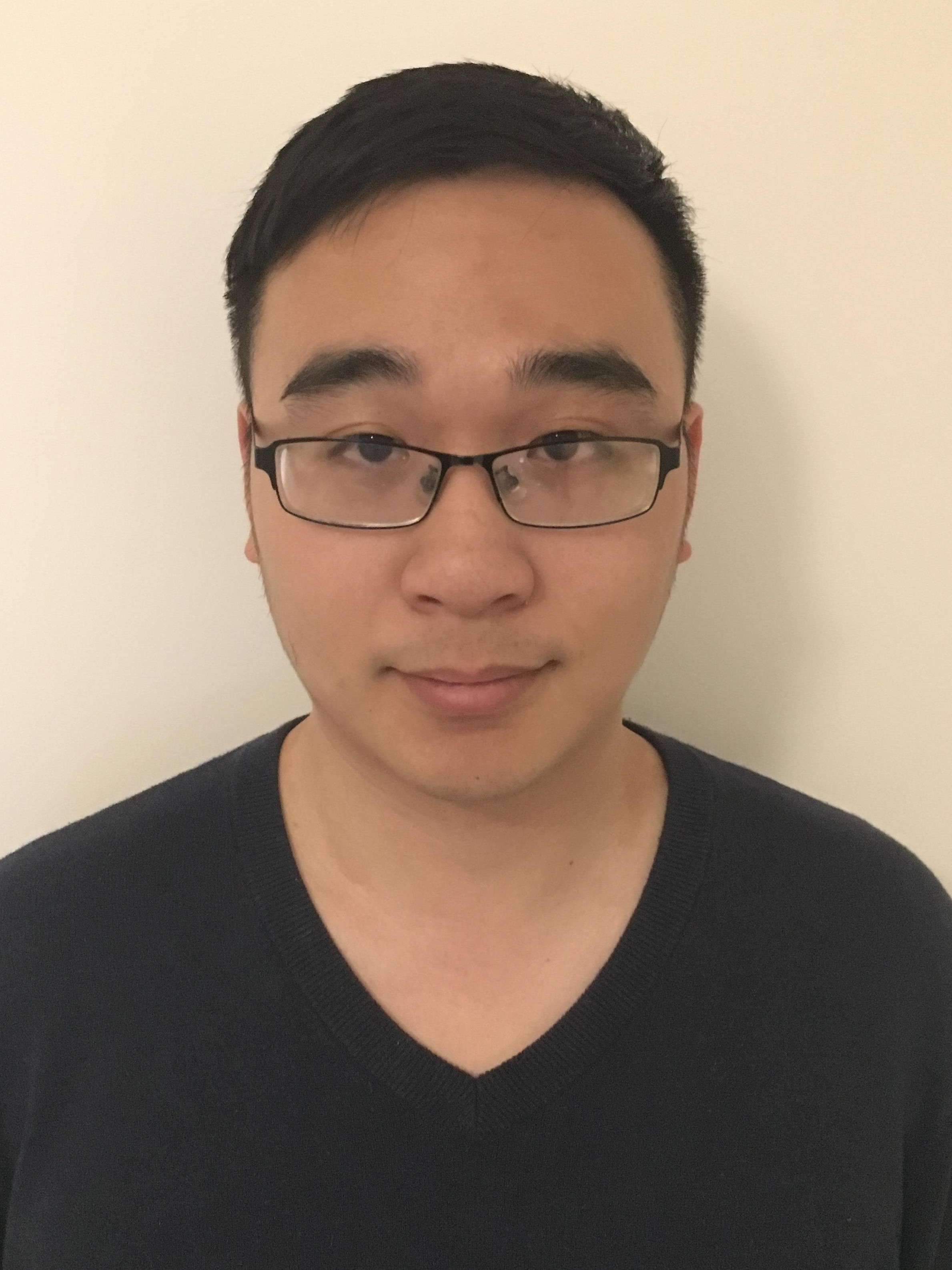}}]{Zheheng Jiang}
received the B.Sc. degree in Electrical Engineering and Automation (Grid Monitoring) from Nanjing Institute of Technology and the M.Sc. degree in Software Development from Queen’s University of Belfast, Belfast, U.K. He has been awarded his Ph.D. degree in Computer Science from University of Leicester, Leicester, U.K. He is currently the Senior Research Associate at the Computing and Communications, Lancaster University, Lancaster, U.K.

His current research interests include machine learning for vision, object detection and recognition, video analysis and event recognition.
\end{IEEEbiography}
\vspace{-3 mm}

\begin{IEEEbiography}[{\includegraphics[width=1in,height=1.25in,clip,keepaspectratio]{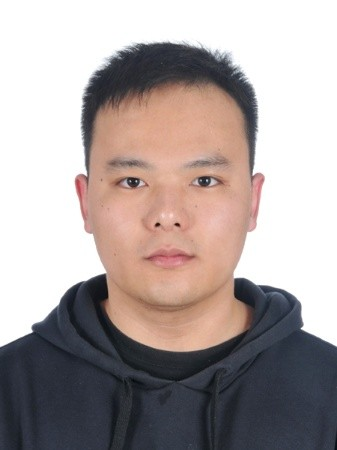}}]{Zhihua Liu} is currently pursuing the Ph.D. degree with the School of Informatics, University of Leicester, Leicester, U.K. His research interests include machine learning, deep learning and computer vision.
\end{IEEEbiography}
\vspace{-3 mm}

\begin{IEEEbiography}[{\includegraphics[width=1in,height=1.25in,clip,keepaspectratio]{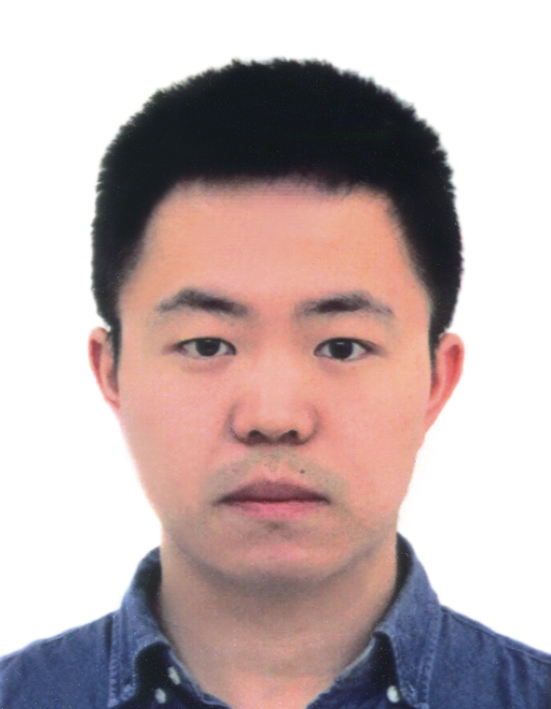}}]{Long Chen} is currently pursuing the PhD degree with the School of Informatics, University of
Leicester, U.K. His research interests are in the areas of Computer Vision and Machine Learning.
\end{IEEEbiography}
\vspace{-3 mm}

\begin{IEEEbiography}[{\includegraphics[width=1in,height=1.25in,clip,keepaspectratio]{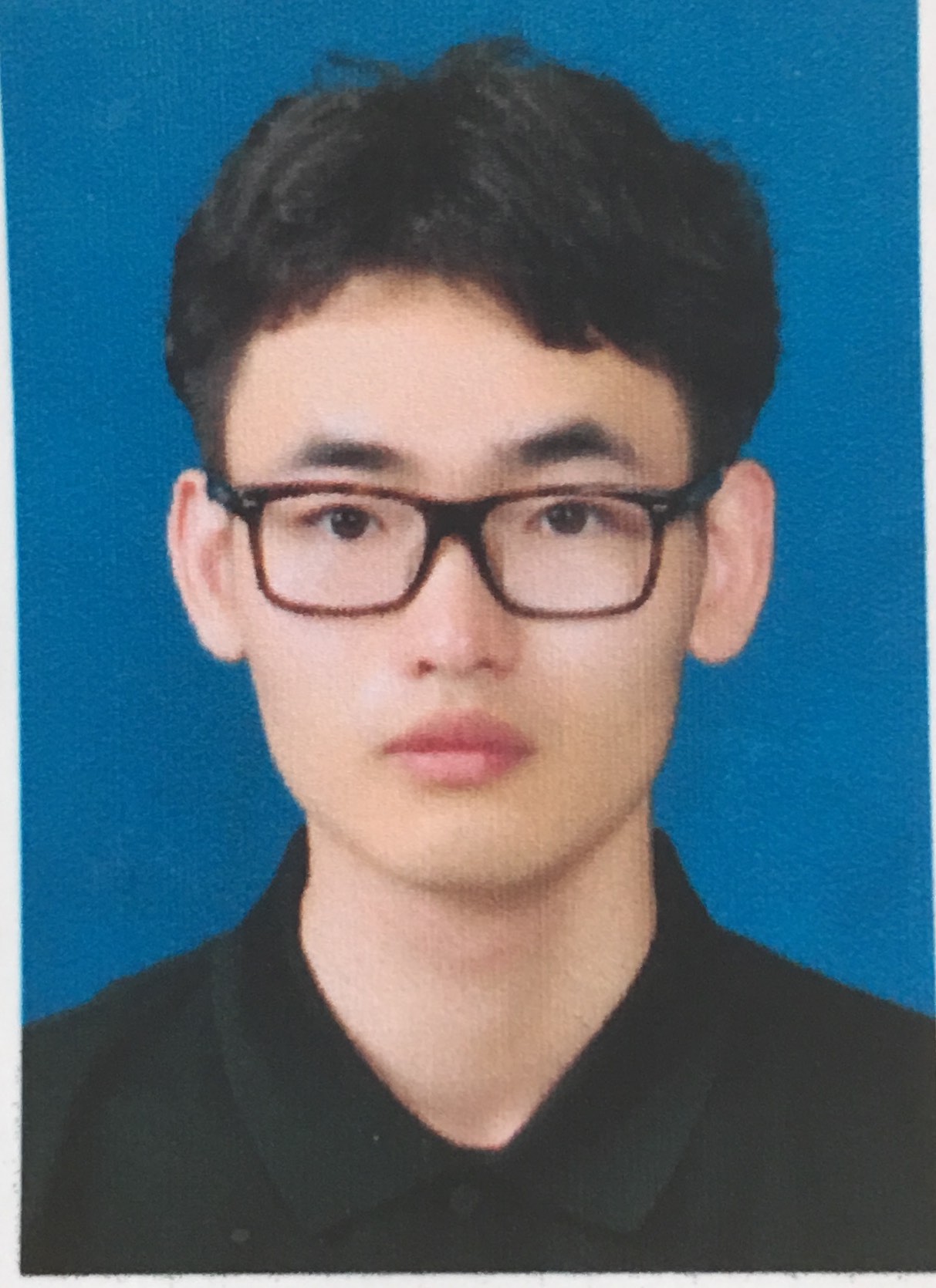}}]{Lei Tong} is currently pursuing the Ph.D. degree with the School of Informatics, University of Leicester, Leicester, U.K. His research interests include computer vision, social network analysis and data mining.
\end{IEEEbiography}
\vspace{-3 mm}

\begin{IEEEbiography}[{\includegraphics[width=1in,height=1.25in,clip,keepaspectratio]{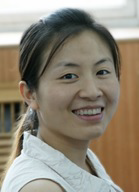}}]{Xiangrong Zhang} received the B.S. and M.S. degrees from the School of Computer Science, Xidian University, Xi’an, China, in 1999 and 2003,
respectively, and the Ph.D. degree from the School
of Electronic Engineering, Xidian University, in
2006. Currently, she is a professor in the Key
Laboratory of Intelligent Perception and Image
Understanding of the Ministry of Education, Xidian University, China. She has been a visiting
scientist in Computer Science and Artificial Intelligence Laboratory, MIT between Jan. 2015 and March 2016. Her research interests include pattern recognition, machine learning, and remote sensing image analysis and understanding.
\end{IEEEbiography}
\vspace{-3 mm}

\begin{IEEEbiography}[{\includegraphics[width=1in,height=1.25in,clip,keepaspectratio]{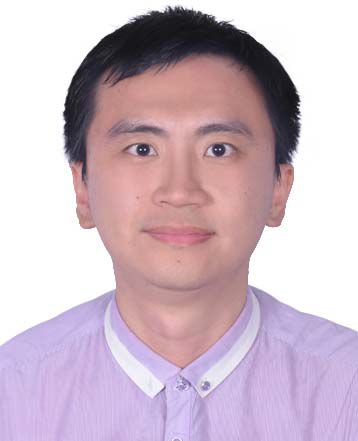}}]{Xiangyuan Lan} received the B.Eng. degree in
computer science and technology from the South
China University of Technology, China, in 2012, and
the Ph.D. degree from the Department of Computer
Science, Hong Kong Baptist University, Hong Kong,
in 2016. He was a Visiting Scholar with the Center
for Automation Research, UMIACS, University of
Maryland at College Park, from January 2015 to
July 2015. His current research interests include
intelligent video surveillance, biometric security, and
health informatics. He is currently an Associate
Editor of Signal, Image and Video Processing (Springer).
\end{IEEEbiography}
\vspace{-3 mm}

\begin{IEEEbiography}[{\includegraphics[width=1in,height=1.25in,clip,keepaspectratio]{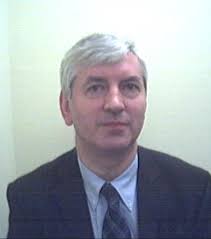}}]{Danny Crookes (SM'12)}
received the B.Sc. and Ph.D. degrees from Queen’s University Belfast,
Belfast, U.K., in 1997 and 1980, respectively. He is currently involved in projects in medical imaging for cancer diagnosis, Speech separation and enhancement, and Capital Markets software using graphics processing units (GPUs).  He was an appointed Professor of Computer Engineeringwith Queen’s University Belfast, in 1993, and was the Head of Computer Science from 1993–2002. His research interests include the use of acceleration technologies (including FPGAs and GPUs) for high performance image, video, and speech processing. He has published some 230 scientific papers in journals and international conferences.
\end{IEEEbiography}
\vspace{-3 mm}

\begin{IEEEbiography}[{\includegraphics[width=1in,height=1.25in,clip,keepaspectratio]{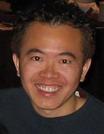}}]{Ming-Hsuan Yang} is a Professor of Electrical Engineering and Computer Science with the University of California at Merced and an Adjunct Professor at Yonsei University. He served as an Associate Editor of the IEEE Transactions on Pattern Analysis and Machine Intelligence from 2007 to 2011, and is an Associate Editor of the International Journal of Computer Vision, Image and Vision Computing, and Journal of Artificial Intelligence Research. He received the NSF CAREER Award in 2012, the Senate Award for Distinguished Early Career Research at UC Merced in 2011, and the Google Faculty Award in 2009. He is a Fellow of the IEEE.

\end{IEEEbiography}
\vspace{-3 mm}

\begin{IEEEbiography}[{\includegraphics[width=1in,height=1.25in,clip,keepaspectratio]{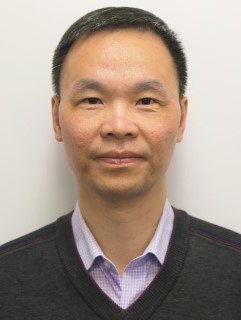}}]{Huiyu Zhou} received a Bachelor of Engineering degree in Radio Technology from Huazhong University of Science and Technology of China, and a Master of Science degree in Biomedical Engineering from University of Dundee of United Kingdom, respectively. He was awarded a Doctor of Philosophy degree in Computer Vision from Heriot-Watt University, Edinburgh, United Kingdom. Dr. Zhou currently is a full Professor at School of Computing and Mathematical Sciences, University of Leicester, United Kingdom. He has published over 400 peer reviewed papers in the field. His research work has been or is being supported by UK EPSRC, ESRC, AHRC, MRC, EU, Royal Society, Leverhulme Trust, Puffin Trust, Invest NI and industry.
\end{IEEEbiography}
\clearpage
\onecolumn

\setcounter{table}{0}
\setcounter{algorithm}{0}
\setcounter{figure}{0}
\setcounter{equation}{0}
\setcounter{page}{1}
\renewcommand\thefigure{S\arabic{figure}}
\renewcommand\thetable{S\arabic{table}}
\renewcommand\thealgorithm{S\arabic{algorithm}}
\section*{Supplementary A}
Table \ref{tab:summary_mouse_tracking} summarizes the mouse based tracking methods described in Section 2.1.
\begin{table}[hbt]
\begin{center}
\caption{Summary of mouse based tracking methods.}
\label{tab:summary_mouse_tracking}
\begin{tabular}{p{2.5cm}p{2.7cm}p{4.5cm}p{3.2cm}p{3cm}}
\toprule
Authors & Requirements & Brief Introduction & Limitation & Part Localization \\
\hline
Twining et al. \cite{twining2001robust}& Standard camera (top view)& Use active shape models to detect targets & Its flexibility is limited because of the required sophisticated skeleton models & Yes (influenced by active shape models)\\
\hline
de Chaumont et al. \cite{de2012computerized}& Standard camera (top view)& Fit the defined geometrical models to images & Its flexibility is limited as it requires sophisticated skeleton models & Yes (influenced by geometrical models)\\
\hline
P{\'e}rez-Escudero et al. \cite{perez2014idtracker}&Standard camera(top view) & Use fingerprints to resolve occlusions and identity combined with motion models & Easily influenced by illumination variation & No\\
\hline
Pistori et al. \cite{pistori2010mice}& Standard camera (top view)& Extension of the standard particle filtering approach & Difficult to explicitly model the entire mouse due to highly deformable shapes & No\\
\hline
Branson et al. \cite{branson2005tracking}& Standard camera (side view) & A particle filtering algorithm for tracking the contours of multiple mice & Difficult to explicitly model the entire mouse due to highly deformable shapes of mice & No\\
\hline
Weissbrod et al. \cite{weissbrod2013automated} & RFID and standard camera (top view) & Use RFID to identify individuals in combination with video data & RFID does not provide sufficient spatial accuracy and temporal resolution & No\\
\hline
Giancardo et al. \cite{giancardo2013automatic} & Thermal camera (top view) & Use thermal camera to detect minor changes in body temperature & Thermal images do not
provide appearance information such as illumination, contrast and texture & Yes (influenced by frame difference and the estimated mouse shape)\\
\hline
Shemesh et al. \cite{shemesh2013high}& Fluorescent colors, UVA light and sensitive color camera (top view) & Lighting change & Marking is invasive and can change mice behaviours & No\\
\hline
Ohayon et al. \cite{ohayon2013automated}& Dye and camera mounted above the enclosure & Dye the fur with different patterns of strokes and dots & Marking is invasive and can modify mice behaviour & Yes (influenced by dye patterns)\\
\hline
Hong et al. \cite{hong2015automated}& Standard camera (top view) and depth camera (top view) & Background subtraction and image segmentation using all cameras & It requires additional equipment and calibration of cameras & Yes (influenced by fitted ellipses)\\
\hline
Sheets et al. \cite{sheets2013quantitative}& Ten synchronised video cameras & Shape-from-silhouette to reconstruct 3D shape of the mouse & It requires additional equipment and calibration of cameras & Yes (influenced by 3D shape)\\
\hline
Romero-Ferrero et al. \cite{romero2019idtracker}& Standard camera (top view) & Use deep-learning-based image classifiers to identify animals that cross paths & This method is to handle occlusions of nearly rigid animals & No\\
\hline
Ours & Standard camera(top view) & Novel BILP Model to associate part candidates with individual targets & Computing time is increasing as the number of mice arises & Yes (by the proposed Bayesian-inference Integer Linear Programming Model)\\
&&\textbf{Advantage: our approach is powerful to track multi-mice parts using a standard camera without color marking or additional equipment.} &&\\
\bottomrule
\end{tabular}
\end{center}
\end{table}

\twocolumn

\section*{Supplementary B}
\begin{algorithm}[H]
\caption{Algorithm for training the proposed PPN.}
\begin{algorithmic}[1]
\renewcommand{\algorithmicrequire}{\textbf{Input:}}
\renewcommand{\algorithmicensure}{\textbf{Output:}}
\REQUIRE a video sequence $\mathcal{F}$ with ground-truth boxes and labels.
\ENSURE  $\left\{\alpha^{z}\right\}_{z=1}^Z$, trained PPN.
 \STATE Initialize the shared convolutional layers by the first 15 layers of VGG-19~ net\cite{simonyan2014very} and all new layers by estimating weights from a zero-mean Gaussian;
 \STATE Create positive and negative root boxes associated with the object $p_{j}\in \left\{ p\right\}_{j=0}^J$ in the first stage;
 \STATE Initially fill the mini-batch of each image with all positive root boxes and randomly sampled negative root boxes;
 \FOR {$z = 1$ to $Z$}
 \STATE Train our PPN $R^{z}$ using Eq. (\ref{equation:loss}) for all objects;
 \STATE Calculate the pseudo-loss of $R^{z}$ using Eq. (\ref{equation:pseudo_loss});
 \STATE Mine hard negative examples by sorting the positive probability of root boxes in $S_{j}^{t}$ of object $p_{j}$;
 \STATE Replace negative examples in previous stage with new hard negative examples;
 \ENDFOR
\STATE Compute $\left\{\alpha^{z}\right\}_{z=1}^Z$ using Eq. (\ref{equation:RPNs_weight});
\RETURN $\left\{\alpha^{z}\right\}_{z=1}^Z$, trained PPN.
\end{algorithmic}
\end{algorithm}

\section*{Supplementary C}
The proof of Eq. (16) is shown below:
As the term $p_{t}\left (X_{k}|a_{n}^{m}\right)$ is of a 2D Gaussian distribution, then we have
\begin{equation}
\begin{split}
\sum_{k\in O}\prod_{m \in M^{*},n\in N^{*}}p_{t}\left (X_{k}|a_{n}^{m}\right)&=\sum_{k\in O}\prod_{m \in M^{*},n\in N^{*}}\mathcal{N}\left (x^{k}_{n,m},\sigma\right)\\&=\sum_{k\in O}\mathcal{N}\left (\left[\sum_{n,m}\frac{x^{k}_{n,m}}{\sigma^2}\right]\frac{\sigma^2}{M*N},\right.\\&\left.\sqrt\frac{\sigma^2}{M*N}\right)\\&=\sum_{k\in O}\mathcal{N}\left (\sum_{n,m}\frac{x^{k}_{n,m}}{M*N},\sqrt\frac{\sigma^2}{M*N}\right)\\&=\mathcal{N}\left (\sum_{k\in O}\sum_{n,m}\frac{x^{k}_{n,m}}{M*N},\right.\\&\left.\sum_{k\in O}\sqrt\frac{\sigma^2}{M*N}\right)
\end{split}
\label{eq:sigma_prod_proof}
\end{equation}

and
\begin{equation}
\begin{split}
\prod_{m \in M^{*},n\in N^{*}}\sum_{k\in O}p_{t}\left (X_{k}|a_{n}^{m}\right)&=\prod_{m \in M^{*},n\in N^{*}}\sum_{k\in O}\mathcal{N}\left (x^{k}_{n,m},\sigma\right)\\&=\prod_{m \in M^{*},n\in N^{*}}\mathcal{N}\left (\sum_{k\in O}x^{k}_{n,m},\sum_{k\in O}\sigma\right)\\&=\mathcal{N}\left (\left[\sum_{n,m}\frac{\sum_{k\in O}x^{k}_{n,m}}{\left(\sum_{k\in O}\sigma\right)^2}\right]\frac{\left(\sum_{k\in O}\sigma\right)^2}{M*N},\right.\\&\left.\sqrt\frac{\left(\sum_{k\in O}\sigma\right)^2}{M*N}\right)\\&=\mathcal{N}\left (\sum_{k\in O}\sum_{n,m}\frac{x^{k}_{n,m}}{M*N},\right.\\&\left.\sum_{k\in O}\sqrt\frac{\sigma^2}{M*N}\right)
\end{split}
\label{eq:sigma_prod_proof2}
\end{equation}
hence:
\begin{equation}
\begin{split}
\sum_{k\in O}\prod_{m \in M^{*},n\in N^{*}}p_{t}\left (X_{k}|a_{n}^{m}\right)=\prod_{m \in M^{*},n\in N^{*}}\sum_{k\in O}p_{t}\left (X_{k}|a_{n}^{m}\right)
\end{split}
\label{eq:sigma_prod_proof3}
\end{equation}

\section*{Supplementary D}
\begin{algorithm}[H]
\caption{Algorithm for optimization of cost function Eq. (17).}
\begin{algorithmic}[1]
\renewcommand{\algorithmicrequire}{\textbf{Input:}}
\renewcommand{\algorithmicensure}{\textbf{Output:}}
\REQUIRE the cost vector $\Phi$, the constraint matrix $A$ and Linear equality constraints $h$.
\ENSURE the optimal solution
\\ \STATE Find the optimal solution to Eq. (17) with the 0-1 restrictions relaxed.
\STATE At the root node, let the relaxed solution be the lower bound $U_{0}$ and randomly select a 0-1 solution with the upper bound $L_{0}$, and set $w=0$.
\WHILE {$U_{w}\neq L_{w}$}
 \STATE Create two new constraints of `$=0$' constraint and `$=1$' constraint for the minimum fractional variable of the optimal solution.
 \STATE Create two new nodes, one for the `$=0$' constraint and one for the `$=1$' constraint.
 \STATE Solve the relaxed linear programming model with the new constraint at each of these nodes.
 \STATE Let the relaxed solution be the lower bound $U_{w+1}$ and the existing maximum 0-1 solution be the upper bound $L_{w+1}$, and set $w=w+1$.
 \STATE Select the node with the minimum lower bound for branching.
\ENDWHILE
\RETURN the optimal solution.
\end{algorithmic}
\label{Alg:optimization}
\end{algorithm}

\section*{Supplementary E}
Alg. \ref{Alg:optimization} is used to seek the global minimum of the cost function Eq. (18) over a $N*\left ( M_{t}+1 \right )$ dimensional solution space $\mathbb{R}^{N*\left ( M_{t}+1 \right )}$. For a subspace $\mathcal{R}\subseteq\mathbb{R}^{N*\left ( M_{t}+1 \right )}$, we define $\Gamma_{min}\left(\mathcal{R}\right)=\min_{\Lambda\in \mathcal{R}}\Phi^T\Lambda$. Thus, $\Gamma_{lb}\left(\mathcal{R}\right)\leq\Gamma_{min}\left(\mathcal{R}\right)\leq\Gamma_{ub}\left(\mathcal{R}\right)$, $\Gamma_{lb}$ and $\Gamma_{ub}$ are functions to compute the lower and upper bounds respectively. We first show that after a large number of iterations $k$, the list of partition $\mathcal{P}_{k}$ must contain a subspace of the original volume. The volume of the subspace is defined as $vol\left(\mathcal{R}\right)=\prod_{i}\left(r_i-l_i\right)$, $\left[l_i,r_i\right]$ is the interval of $\mathcal{R}$ along the $i$th dimension and therefore
\begin{equation}
\min_{\mathcal{R}\in\mathcal{P}_{k}}vol\left(\mathcal{R}\right)\leq\frac{vol\left(\mathbb{R}^{N*\left ( M_{t}+1 \right )}\right)}{k}
\label{eq:min_vol}
\end{equation}
We also have
\begin{equation}
vol\left(\mathcal{R}\right)\geq \max_{i}\left(r_i-l_i\right)\left(\min_{i}\left(r_i-l_i\right)\right)^{N*\left ( M_{t}+1 \right )-1}
\label{eq:vol}
\end{equation}
We define the condition number of subspace $\mathcal{R}$ as follows:
\begin{equation}
cond\left(\mathcal{R}\right)=\frac{\max_{i}\left(r_i-l_i\right)}{min_{i}\left(r_i-l_i\right)}
\label{eq:cond}
\end{equation}
Since we define $\mathcal{R}$ as the minimum lower bound as described in Alg.\ref{Alg:optimization}, we have
\begin{equation}
cond\left(\mathcal{R}\right)\leq \max\left\{cond\left(\mathbb{R}^{N*\left ( M_{t}+1 \right )},2\right)\right\}
\label{eq:cond_max}
\end{equation}
Combing Eqs. (\ref{eq:min_vol}), (\ref{eq:vol}), (\ref{eq:cond}) and (\ref{eq:cond_max}), we have:
\begin{align}
\min_{\mathcal{R}\in\mathcal{P}_{k}}length\left(\mathcal{R}\right)\leq \max\left\{cond\left(\mathbb{R}^{N*\left ( M_{t}+1 \right )},2\right)\right\}\nonumber\\\left(\frac{vol\left(\mathbb{R}^{N*\left ( M_{t}+1 \right )}\right)}{k}\right)^{\frac{1}{N*\left ( M_{t}+1\right )}}
\label{eq:min_length}
\end{align}
where $length(\mathcal{R})=\max_{i}\left(r_i-l_i\right)$. Thus, with the increasing  $k$, the maximum dimension of $\mathcal{R}_{min}$, which is the smallest subspace in $\mathcal{P}_k$, is decreasing. As $length(\mathcal{R})$ goes to zero, the difference between the upper and lower bounds (i.e., $\Gamma_{ub}\left(\mathcal{R}\right)-\Gamma_{lb}\left(\mathcal{R}\right)$) uniformly converges to zero.

\section*{Supplementary F}
\begin{figure}
\begin{center}
\includegraphics[width=9cm]{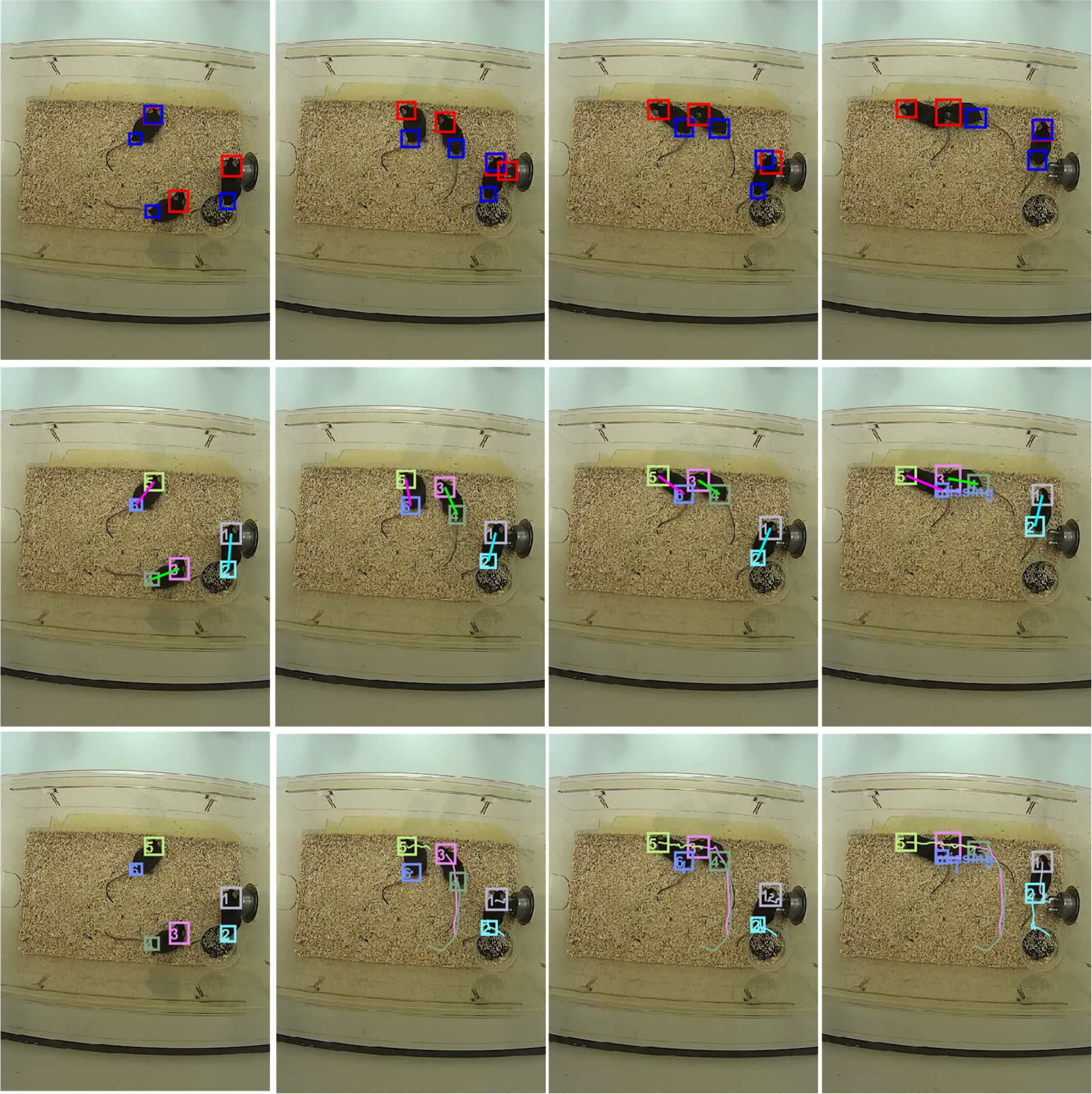}
\end{center}
\caption{\textbf{Top:} Part (mouse head and tail) detection candidates shown over four frames. \textbf{Middle:} Estimated locations of the parts for all the mice. Each colourful line corresponds to a unique mouse identity and each colourful bounding box corresponds to a unique part identity. \textbf{Bottom}: Estimated trajectories of all the parts.}
\label{fig:tracking_parts}
\end{figure}

Fig. \ref{fig:tracking_parts} shows the part tracking of multi-mice described in Section 4.3.

\begin{table}
\begin{center}
\caption{Comparison with conventional methods on the 2-Mice dataset.}
\label{tab:tracking__convention_2m}
\begin{tabular}{p{2cm}p{1.2cm}p{1.2cm}p{1.2cm}p{1.2cm}}
\toprule
Method & MOTA$\%\uparrow$ & MOTP$\%\uparrow$  & IDF1$\%\uparrow$ & FPS$\uparrow$\\
\hline
Motr \cite{ohayon2013automated}& 62 & 53.9  & 69 & 30\\

MiceProfiler \cite{de2012computerized}& 49.6 & 56.0  & 28.6 & 26\\

our & 81.1 & 65.5  & 91.2 & 30.8\\

\bottomrule
\end{tabular}
\end{center}
\end{table}

\begin{figure}
\begin{center}
\begin{tabular}{cc}
\includegraphics[width=3.2cm]{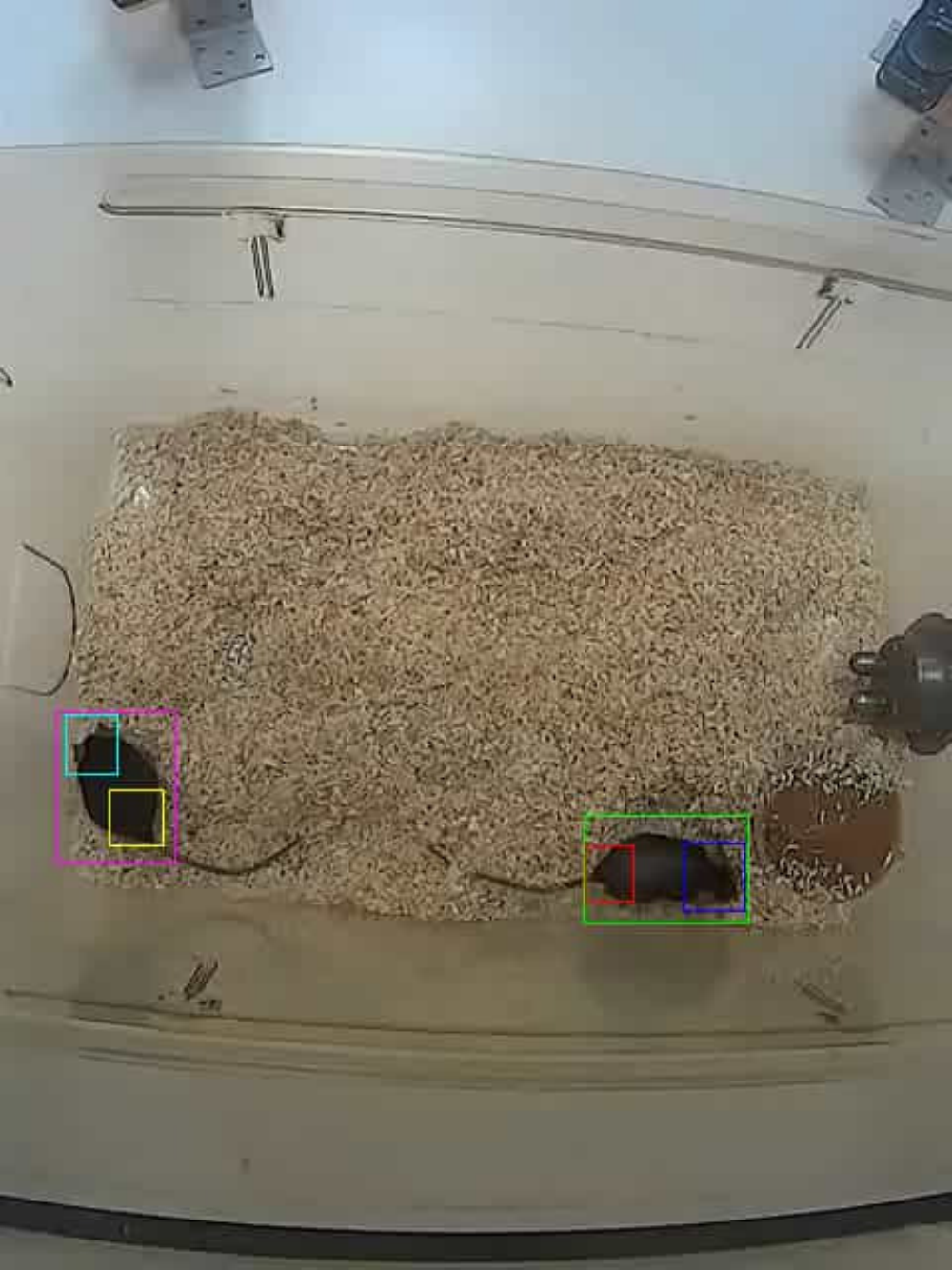}&
\includegraphics[width=3.2cm]{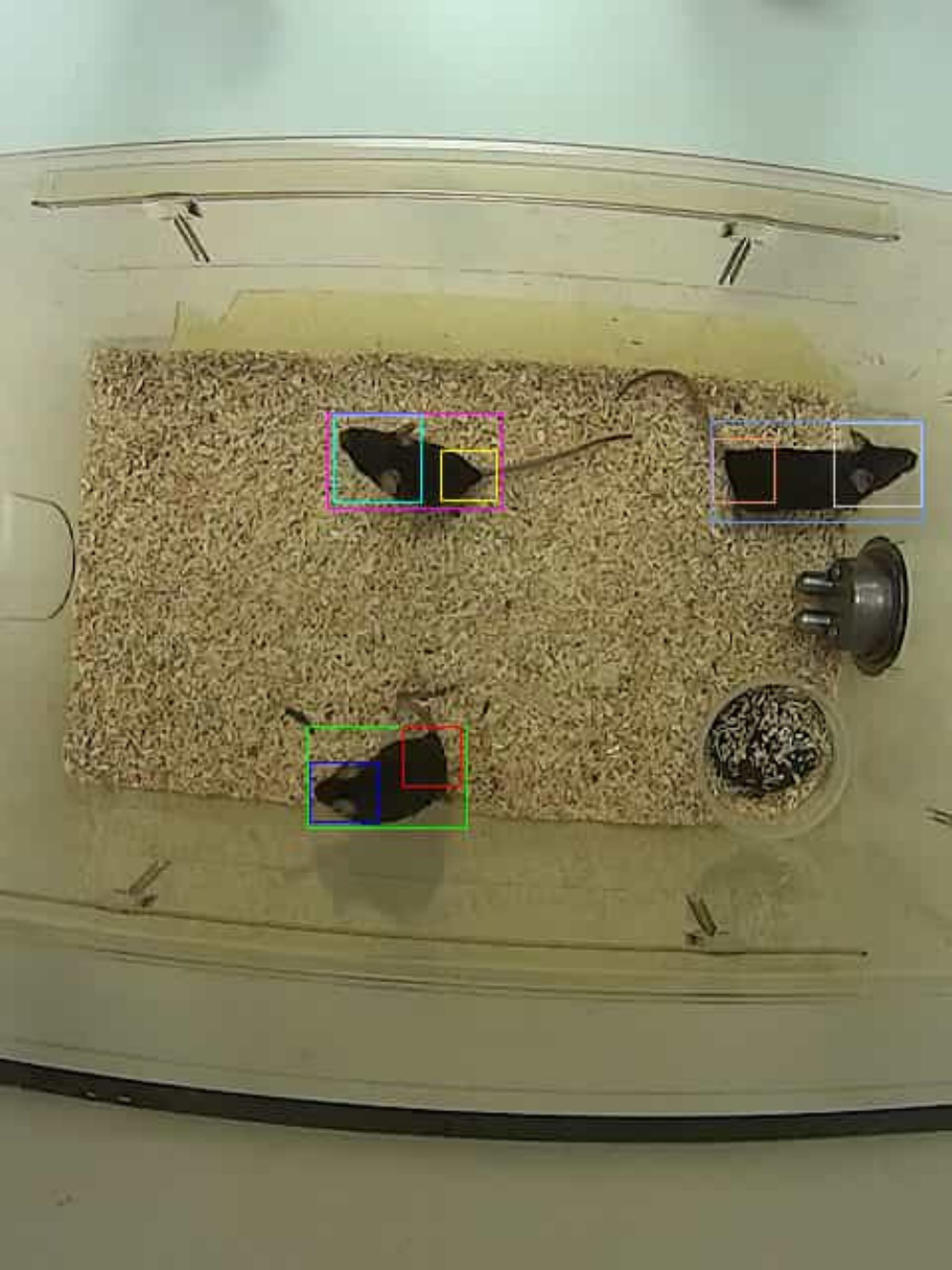}\\
\end{tabular}
\end{center}
\caption{Exemplar frames and annotations from the proposed Multi-Mice PartsTrack dataset.}
\label{fig:database}
\end{figure}

Fig. \ref{fig:database} shows some exemplar frames and annotations from the proposed Multi-Mice PartsTrack dataset described in Section 5.1.

\begin{figure*}
\begin{center}
\begin{tabular}{ccc}
\includegraphics[width=5cm]{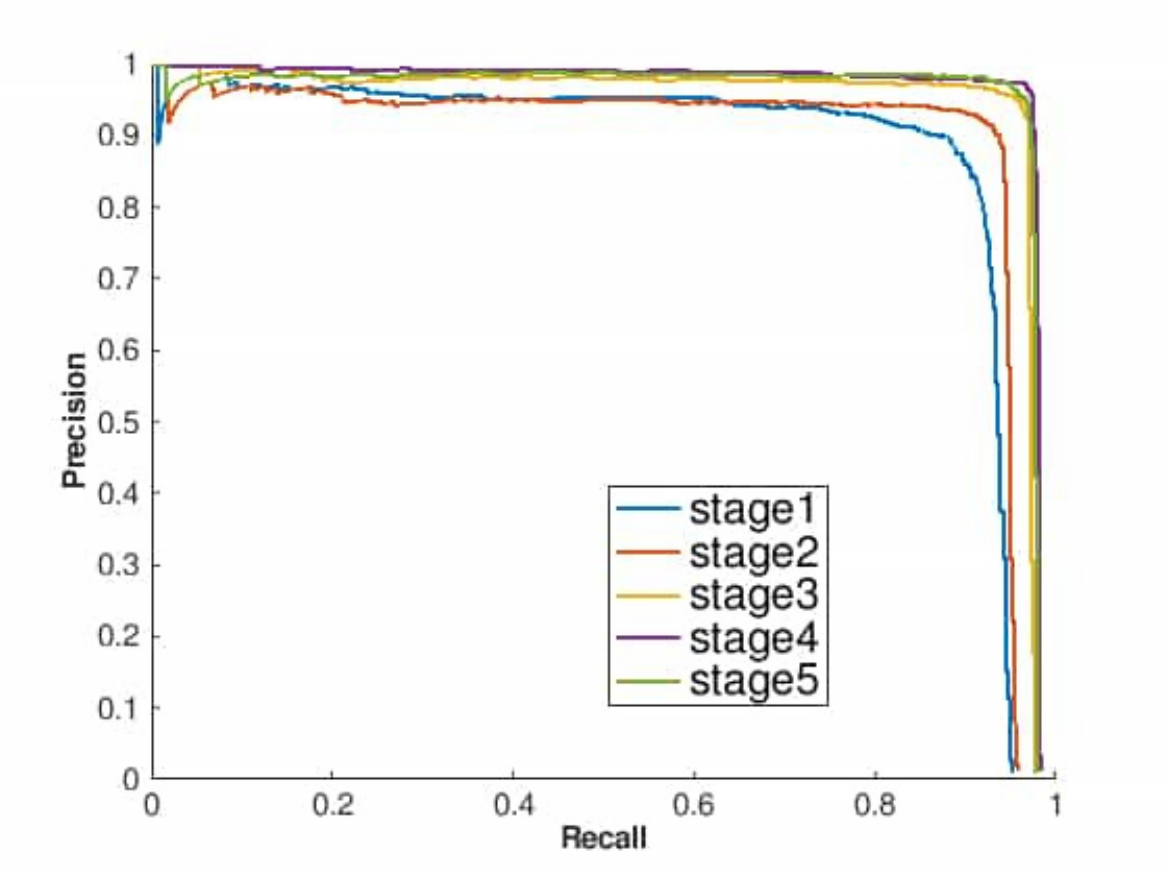}&
\includegraphics[width=5cm]{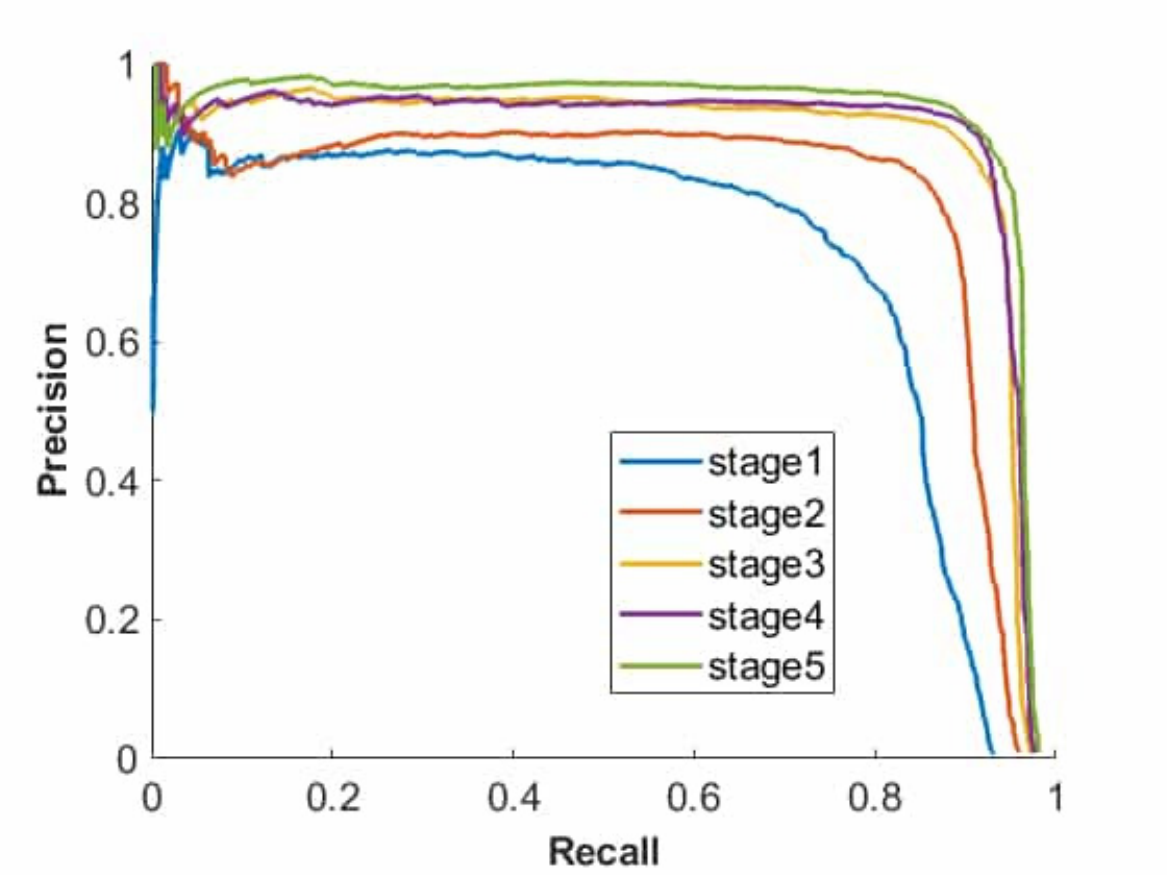}&
\includegraphics[width=5cm]{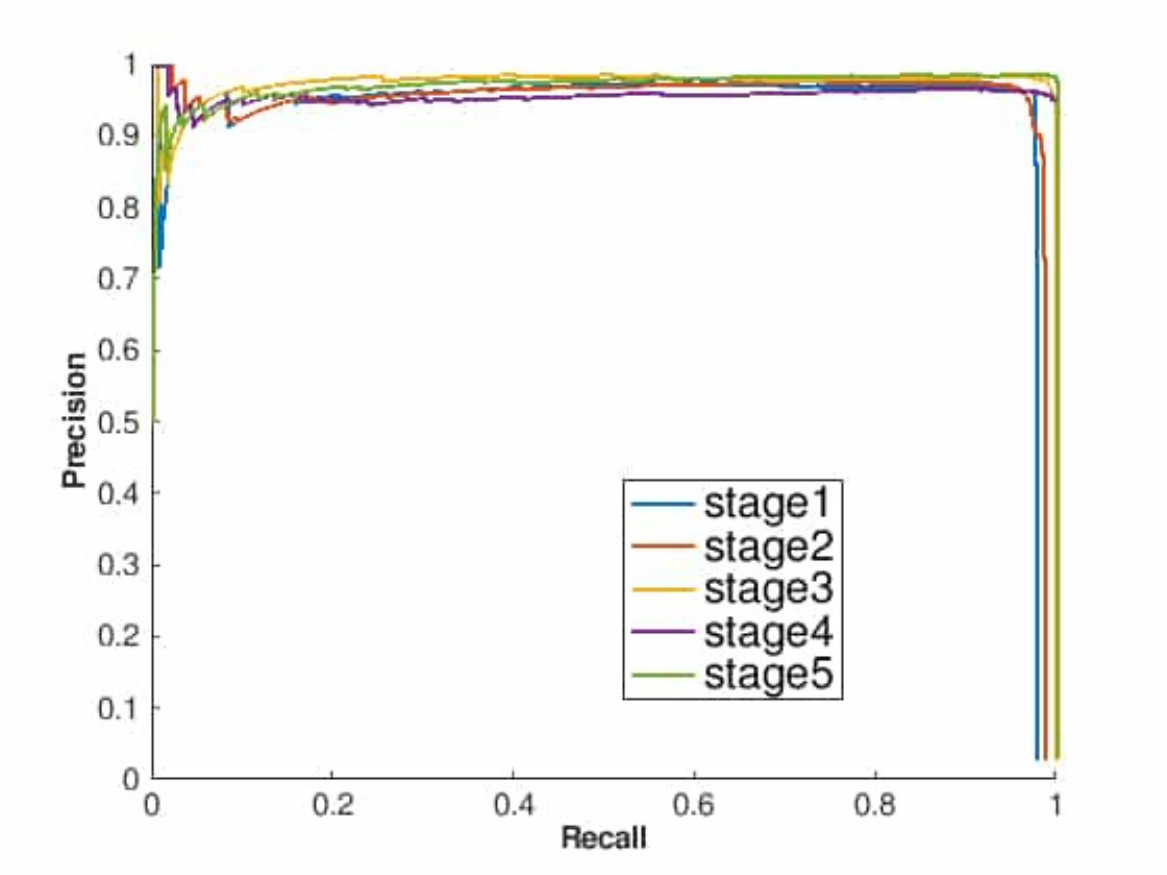} \\
(a) head&(b) tail&(c) body
\end{tabular}
\end{center}
\caption{Precision/Recall curves of the proposed Part and Body Proposal Network across various stages on the 2-mice dataset.}
\label{fig:stage}
\end{figure*}

\begin{table*}
\begin{center}
\caption{Quantitative evaluation of multi-mice parts tracking on the 2-Mice dataset.}
\label{tab:tracking_results_2m}
\begin{tabular}{p{4cm}p{1.2cm}p{1.2cm}p{1.2cm}p{1.2cm}p{1.2cm}p{1.2cm}p{1.2cm}p{1.2cm}}
\toprule
Method & MOTA & MOTP &  MT & ML & FP & FN & IDs & IDF1\\
& $\%\uparrow$ & $\%\uparrow$ & $\uparrow$ &$\downarrow$&$\downarrow$&$\downarrow$&$\downarrow$&$\%\uparrow$\\
\hline
\multicolumn{8}{c}{\textit{Impact of the constraints}}\\
\hline
All  & \textbf{85.0} & 65.5  & \textbf{19} & \textbf{0} & \textbf{272} & \textbf{288} & \textbf{8} & \textbf{86.6}\\
All(b,c,d)  & 84.6 & 65.4  & 19 & 0 & 280 & 288 & 18 & 83.0\\
All(a,c,d)  & 81.0 & 66.1  & 17 & 2 & 374 & 331 & 18 & 80.9\\
All(a,b,d)  &82.7 & 66.0 & 18  & 0 & 342 & 299 & 14 &  82.4 \\
All(a,b,c)  & 80.1 & 69.6  & 17 &  0 & 399 & 356 & 18 & 79.8\\
\hline
\multicolumn{8}{c}{\textit{Impact of the geometric model}}\\
\hline
All without geometric model & 71.4 & 66.1  & 15 & 0 & 557 & 514 & 15 & 65.5\\
\hline
\multicolumn{8}{c}{\textit{Impact of the motion model}}\\
\hline
All without motion model & 70.1 & 65.4  & 17 & 2 & 576 & 533 & 27 & 67.4\\
\hline
\multicolumn{8}{c}{\textit{Impact of the parts association model}}\\
\hline
All without parts association model & 74.4 & 66.3  & 16 & 0 & 487 & 462 & 41 & 71.7\\
\hline
\multicolumn{8}{c}{\textit{Comparison with the other state-of-the-arts}}\\
\hline
MOTDT [77] & 22.2 & 30.4  & 7 & 0 & 658 & 504 & 830 & 40.4\\

MDP [39] & 67.9 & 69.9  & 14 & 0 & 630 & 583 & 10 & /\\

SORT [40] & 56.7 & 69.5  & 7 & 0 & 571 & 509 & 10 & 32.2\\

JPDAm [45] & 55.0 & 67.6  & 11 & 0 & 858 & 815 & 34 & 60.0\\

RNN-LSTM~\cite{milan2017online} & 41.2 & 62.6  & 9 & 0 & 983 & 916 & 22 & 63.8\\
CEM~\cite{milan2013continuous} & 37.4 & 68.5  & 9 & 0 & 1138 & 904 &  47 & /\\
\bottomrule
\end{tabular}
\end{center}
\end{table*}

In Fig. \ref{fig:stage}, we conduct ablation study on the multi-stage training design and illustrate the Precision/Recall curves of the proposed Part and Body Proposal Network across various stages on the 2-mice dataset described in Section 6.1.2.

\begin{figure}
\begin{center}
\begin{tabular}{cc}
\includegraphics[width=3.5cm]{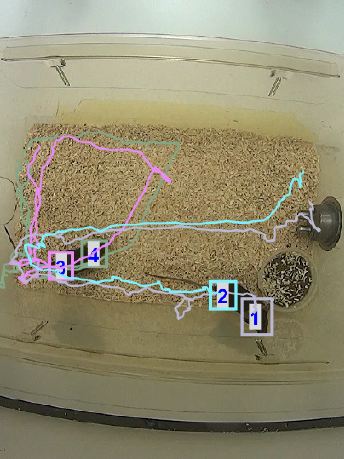}&
\includegraphics[width=3.5cm]{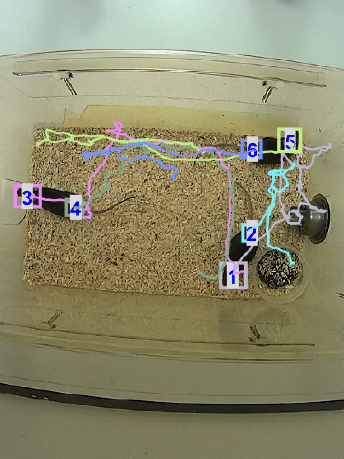}\\
\includegraphics[width=3.5cm]{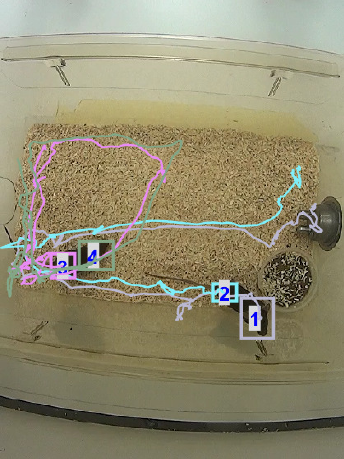}&
\includegraphics[width=3.5cm]{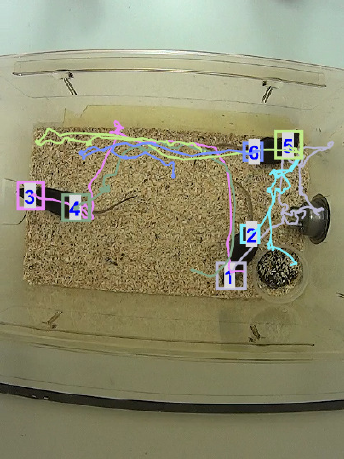}\\
(a)&(b)
\end{tabular}
\end{center}
\caption{Tracking ground-truth (top) and results (bottom) of the test sequences in the proposed Multi-Mice PartsTrack dataset. The trajectory and rectangle of each mouse part are shown in different colours: (a) The leftmost mouse: head (labelled as 4) is green and tail (labelled as 3) is pink, the rightmost mouse: head (labelled as 1) is grey and tail (labelled as 2) is cyan; (b) the leftmost mouse: head (labelled as 3) is pink and tail (labelled as 4) is green, the upper right mouse: head (labelled as 5) is GreenYellow and tail(labelled as 6) is blue, the lower right mouse: head (labelled as 1) is grey and tail (labelled as 2) is cyan.}
\label{fig:tracking_results}
\end{figure}

Tab. \ref{tab:tracking_results_2m} shows quantitative evaluation of multi-mice part tracking on the 2-Mice dataset described in Section 6.2.2. Fig. \ref{fig:tracking_results} shows exemplar tracking results of the test sequences in the proposed Multi-Mice PartsTrack dataset. 

\begin{figure}
\begin{center}
\begin{tabular}{cc}
\includegraphics[width=3.5cm]{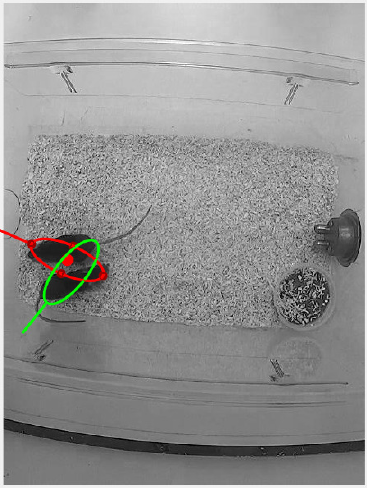}&
\includegraphics[width=3.5cm]{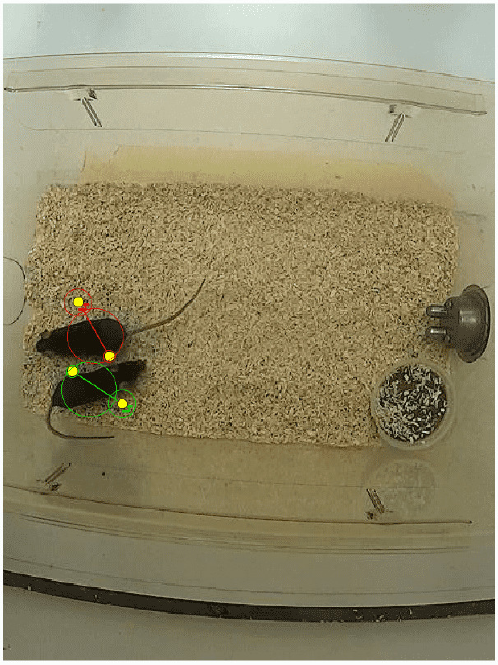}\\
Motr\cite{ohayon2013automated} & MiceProfiler\cite{de2012computerized}\\
\includegraphics[width=3.5cm]{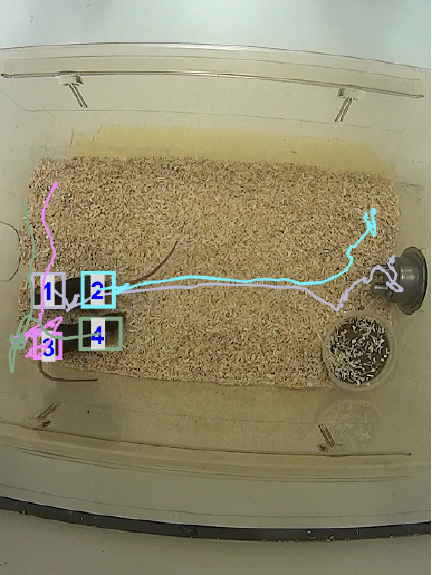}&
\includegraphics[width=3.5cm]{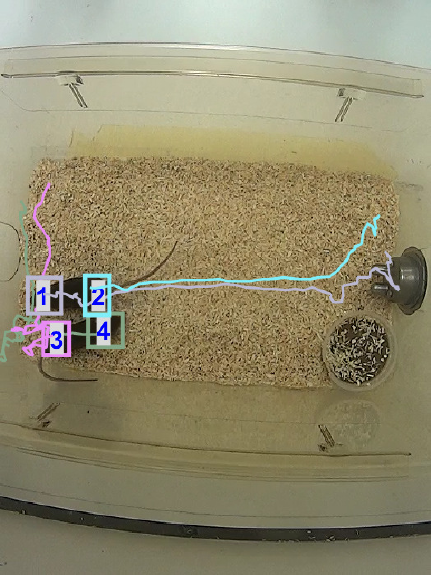}\\
Ours & ground truth\\
\end{tabular}
\end{center}
\caption{Comparison with conventional methods i.e., Motr and MiceProfiler. Motr and MiceProfiler track mice by fitting ellipses and hard-coded geometric models. When the mice are very close, the models used in these methods cannot be properly fitted, leading to tracking targets’ swaps or incorrect parts localization.}
\label{fig:tracking_convention}
\end{figure}

\begin{figure*}[hbt]
\begin{center}
\includegraphics[width=18cm]{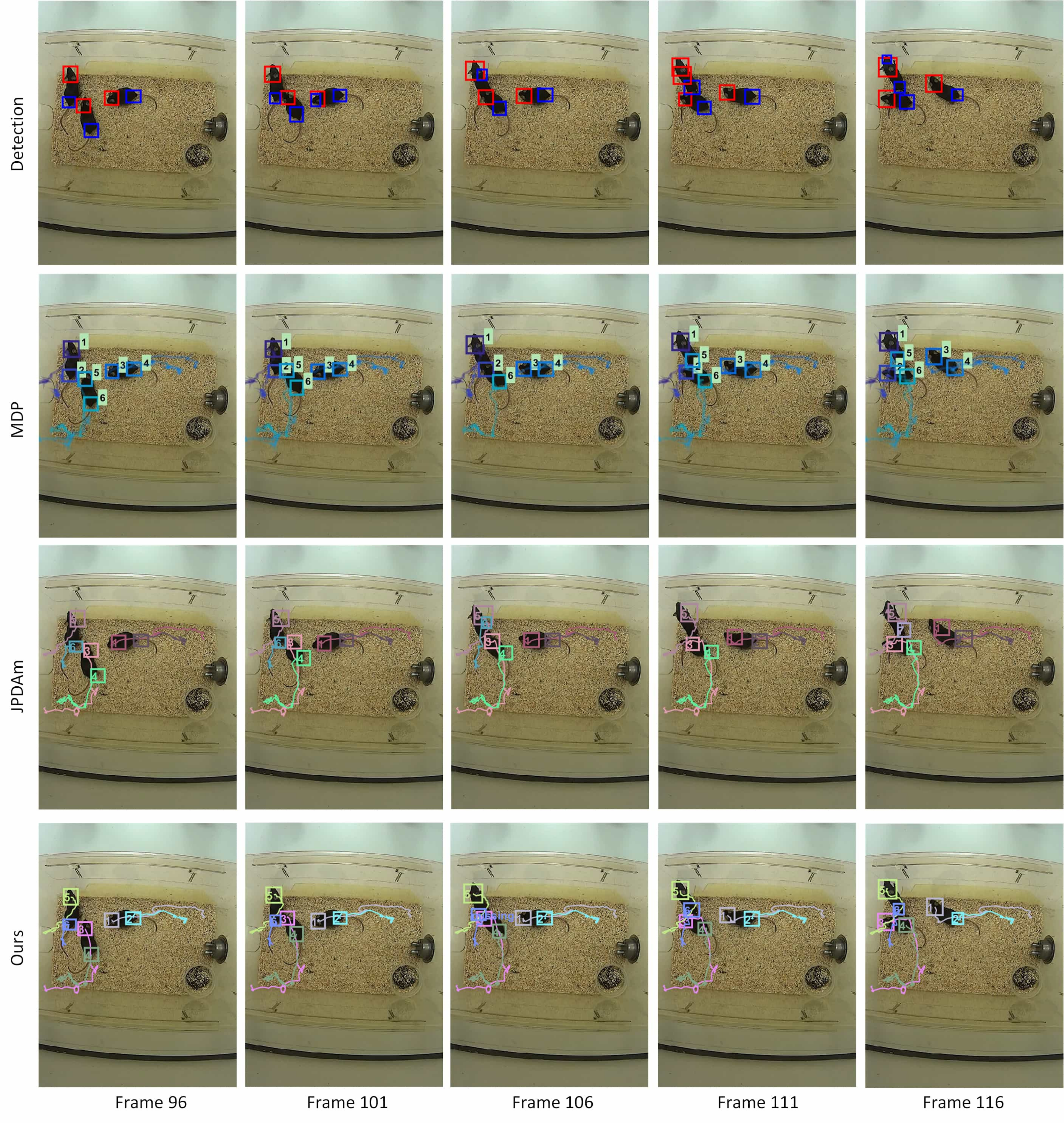}
\end{center}
\caption{Qualitative comparison of the proposed tracking method (row 4) against MDP \cite{xiang2015learning} (row 2) and JPDAm \cite{hamid2015joint} (row 3) using the same detection results (row 1). MDP swap identities between targets 2 and 5 after the occlusion occurs due to `Pinning' at the frame 106, i.e., the target identity is swapped between the tail of the upper left mouse and the head of the lower left mouse. JPDAm assigns a new identity to target 6 after it is occluded by target 3, i.e., the occluded tail of the upper left mouse is assigned to a new identity number. }
\label{fig:tracking_example1}
\end{figure*}

\begin{figure*}[hbt]
\begin{center}
\includegraphics[width=18cm]{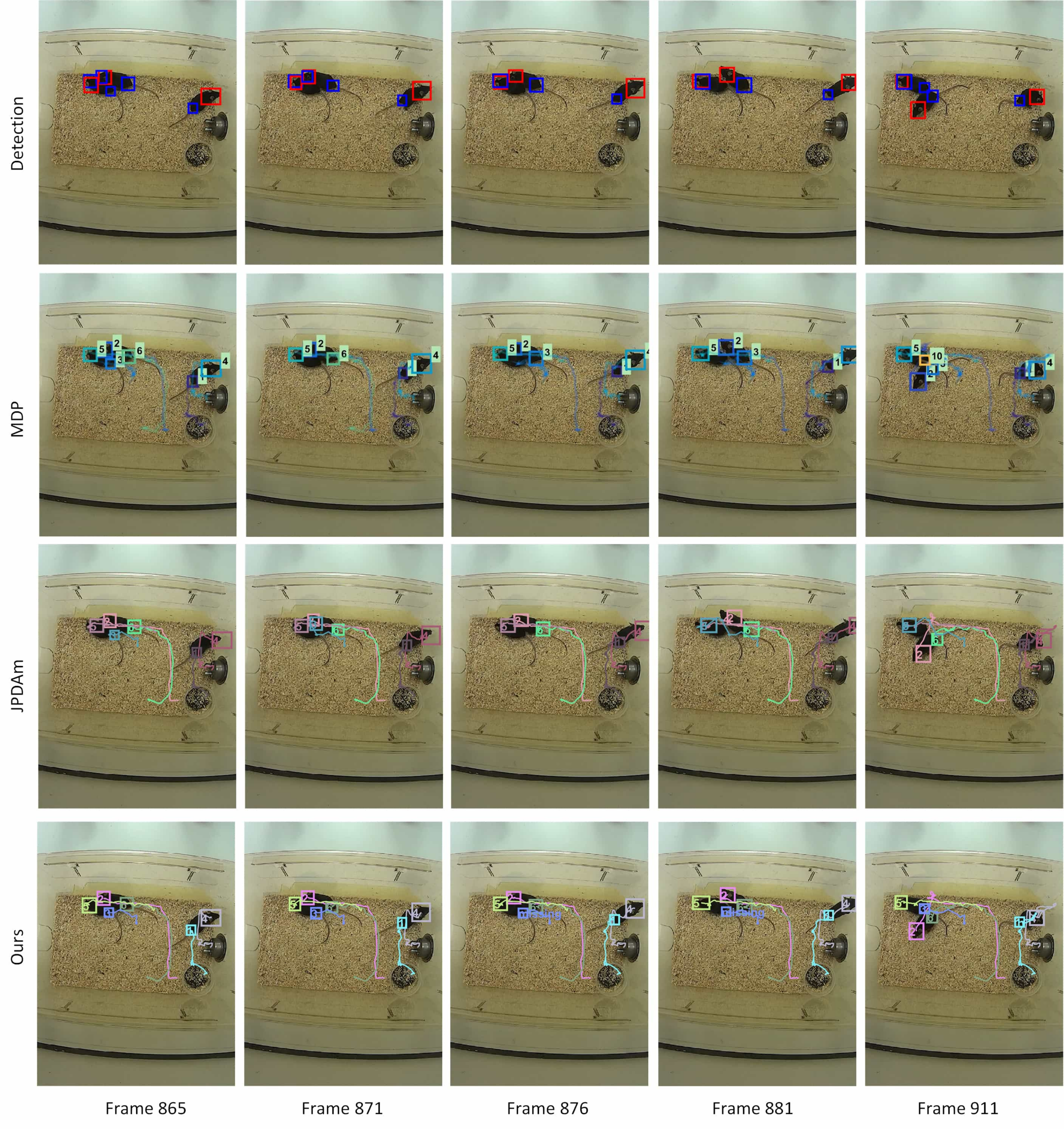}
\end{center}
\caption{Qualitative comparison of the proposed tracking method (row 4) against MDP \cite{xiang2015learning} (row 2) and JPDAm \cite{hamid2015joint} (row 3) using the same detection results (row 1). As shown at image frame 876 (for the MDP algorithm), target 3 (the tail of the leftmost mouse) sees drifts and switches to target 6 (the tail of the middle mouse). After the drifting of target 3, the original object is assigned to a new identity number as shown at image frame 911. In the JPDAm algorithm, target 3 (the tail of the leftmost mouse) drifts towards target 5 (the head of the leftmost mouse) and finally switches to target 5 at image frame 911.}
\label{fig:tracking_example3}
\end{figure*}

\begin{figure*}[hbt]
\begin{center}
\includegraphics[width=18cm]{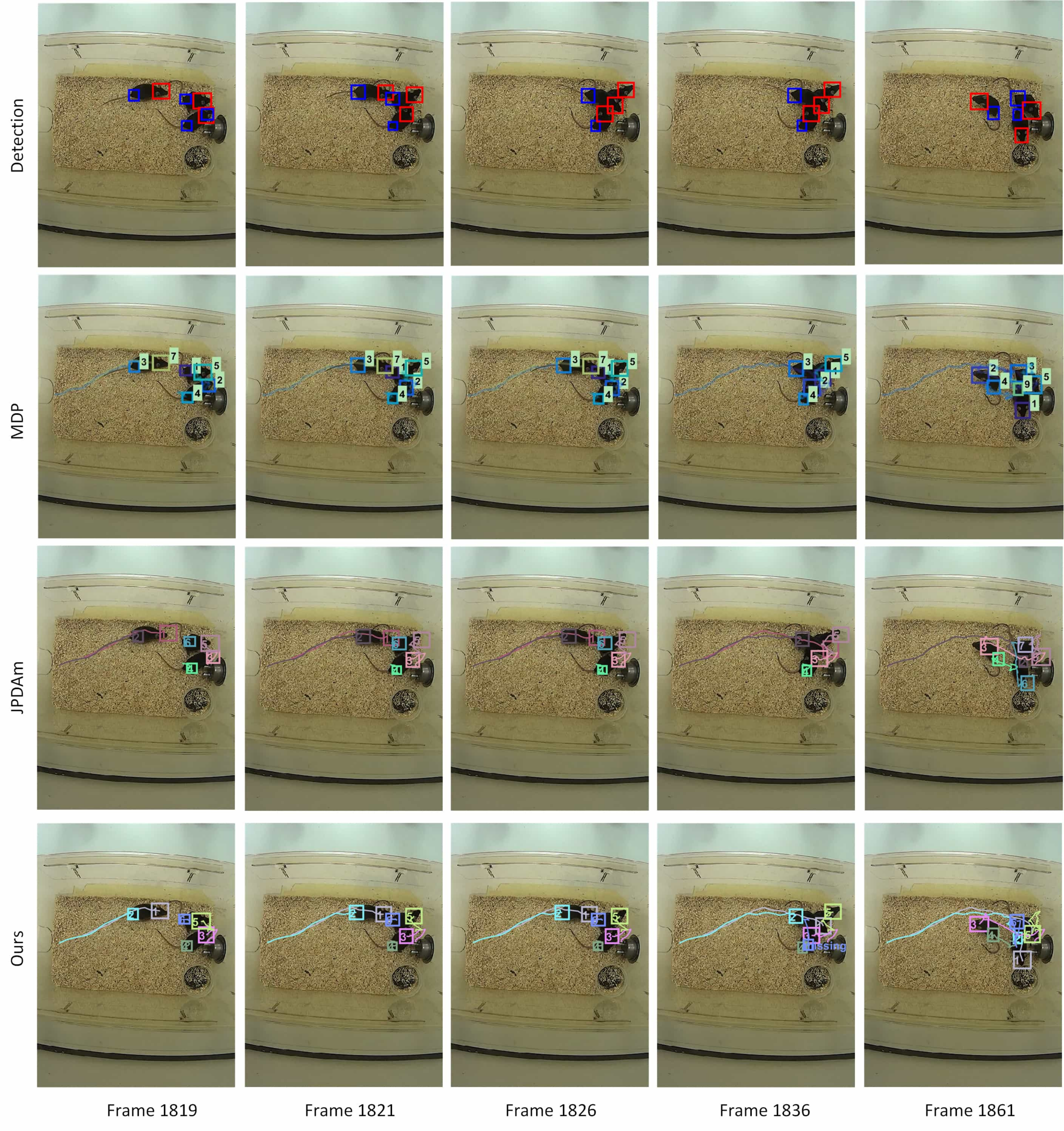}
\end{center}
\caption{Qualitative comparison of the proposed tracking method (row 4) against MDP \cite{xiang2015learning}(row 2) and JPDAm \cite{hamid2015joint} (row 3) using the same detection results (row 1). The problems of identity swap and target drift are more serious in this situation. In the MDP algorithm, target 3 (the tail of the leftmost mouse at image frame 1819) has drifts and switches to target 1 (the tail of the upper right mouse in the first column) at image frame 1861. Moreover, in the same frame, target 7 (the head of the leftmost mouse at image frame 1819) is replaced by target 1, and its tail is assigned to a new identity number 9. Similar problems occur for JPDAm, where target 6 (the tail of the upper right mouse at image frame 1819) replaces target 1 (the head of the upper right mouse at image frame 1819) and target 6 is replaced by a new identity number 7 at image frame 1861. Although target 6 (the tail of the upper right mouse at image frame 1819) in our algorithm also causes a drift due to occlusion at image frame 1836, target 6 finally finds the correct object when the occluded part appear again at image frame 1861.}
\label{fig:tracking_example4}
\end{figure*}

Fig. \ref{fig:tracking_convention} show tracking results of conventional methods (Motr and MiceProfiler). Figs. \ref{fig:tracking_example1}, \ref{fig:tracking_example3} and \ref{fig:tracking_example4} provide the qualitative comparison of the proposed tracking method against MDP [49] and JPDAm [63] described in Section 6.2.2.

\begin{figure*}[hbt]
\begin{center}
\includegraphics[width=12cm]{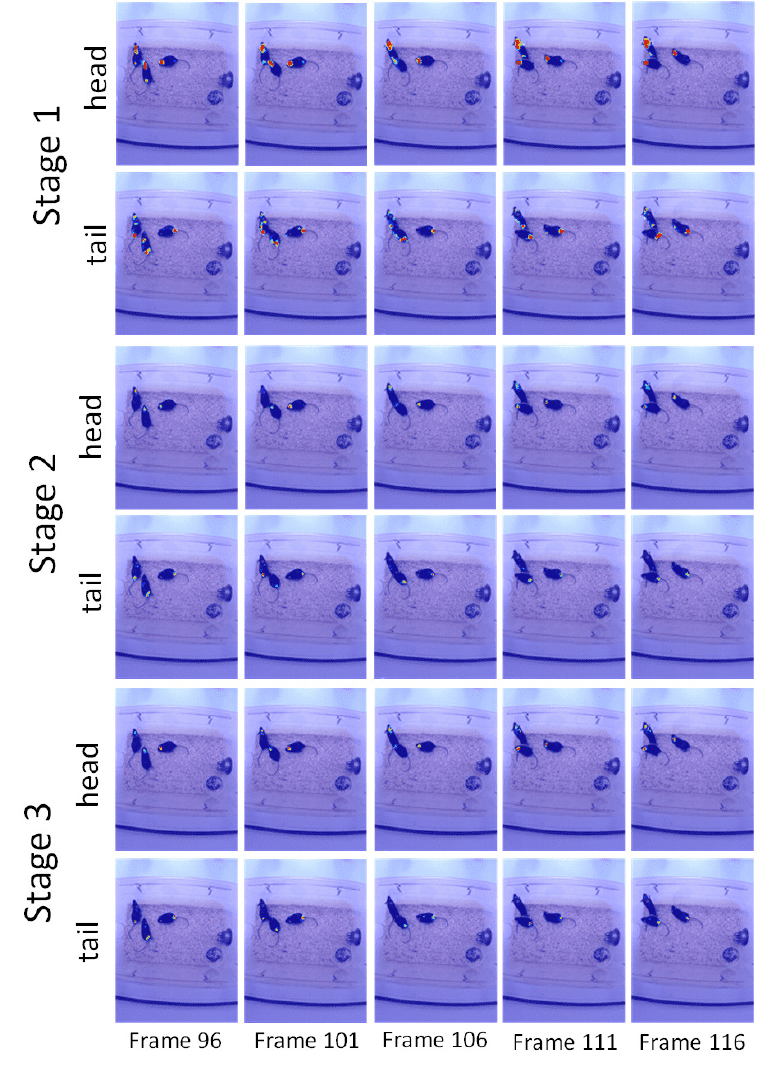}
\end{center}
\caption{Confidence maps of the mouse head and tail bases in Fig. \ref{fig:tracking_example1}.}
\label{fig:feat_map}
\end{figure*}

\end{document}